%% file: main.tex
\PassOptionsToPackage{dvipsnames}{xcolor}
\documentclass[11pt]{article}
\usepackage[hyperref]{acl}
\usepackage{times}
\usepackage{latexsym}
\usepackage{graphicx}
\usepackage{fancyvrb} % BVerbatim: a boxable verbatim we can \resizebox to fit the margins
\usepackage{booktabs}
\usepackage{amsmath}
\usepackage{amssymb}
\usepackage{multirow}
\usepackage{enumitem}
\usepackage{placeins}
\usepackage{tikz}
\usetikzlibrary{arrows.meta, positioning, fit, backgrounds, calc, shapes.geometric}

\newcommand{\taubench}{$\tau$2-bench}
\newsavebox{\judgepromptbox} % holds the Fig. 9 judge prompt so it can be scaled to \textwidth
\newcommand{\OurMethod}{PROOF-Gen}
\newcommand{\OurMethodFull}{Per-scenario Reflective Optimization to Overcome Failed Generation}

\title{\OurMethod{}: From Optimized Data to Better Distillation}

\author{
  Anh Ta \quad Junjie Zhu \quad Shahin Shayandeh \\
  Apple \\
  \texttt{\{atta, jason.zhu, shn\}@apple.com}
}

\begin{document}
\maketitle

\begin{abstract}
\input{sections/abstract}
\end{abstract}

\input{sections/introduction}

\input{sections/method}
\input{sections/experimental_setup}
\input{sections/results}

\input{sections/related_work}
\input{sections/conclusion}

% Back matter. None of these count toward the 7-page camera-ready content
% limit; Limitations must precede the references (required by the call).
\input{sections/limitations}
\input{sections/ethics}
\input{sections/acknowledgments}

\bibliography{references}

\input{sections/appendix}

\end{document}

%% file: sections/abstract.tex
Supervised fine-tuning on teacher-generated trajectories is the standard first stage for distilling tool-calling capabilities into deployable models. Post-training pipelines that drive shipped tool-calling agents re-run this stage on a daily or weekly cadence, paying the frontier-teacher cost each cycle, yet the mechanism is generate-and-filter (keep the teacher's passing trajectories, discard the rest) and each cycle leaves behind the same hard scenarios because failures supply no signal. On \taubench{}, 57\% of teacher trials fail, two-thirds of them \emph{near-misses} (most tool calls correct, undone by one decisive error).

We introduce \OurMethod{} (\OurMethodFull{}), which recovers golden trajectories from these failures via per-scenario prompt optimization. For each failed task, a reflector analyzes the execution trace and evaluation feedback, then writes corrective guidance that steers the teacher to a passing trajectory. The guidance is stripped before training, so the student learns from clean demonstrations with no task-specific scaffold.

On \taubench{}, per-scenario optimization recovers 93\% of failed scenarios. Fine-tuned on the combined data, Qwen3-4B-Instruct-2507 improves from Pass\textasciicircum{}1=0.132 to 0.529 and Gemma 4 E4B-it gains +7.2pp on BFCL v4 multi-turn. In a deployed pipeline, the method lifts trajectory quality by +6.3pp goal completion and transfers to a deployed on-device model (+1.5pp goal completion; +1.7 to +5.0pp across response-quality metrics), with positive transfer in every locale (non-English average +1.48pp).

%% file: sections/introduction.tex
\section{Introduction}
\label{sec:introduction}

Distillation via supervised fine-tuning (SFT) is the dominant first
stage for transferring capabilities from frontier models to the
smaller, deployable models that drive tool-calling agents in production~\cite{ICLR2024_5be69a58,li2025naturalthoughtsselectingdistillingreasoning,luo2026agentarkdistillingmultiagentintelligence}.
Post-training cycles run on a daily or weekly cadence to absorb new
scenarios, tools, and policies; every cycle pays the same
frontier-teacher cost, and every cycle leaves behind the same hard
scenarios, because the standard generate-and-filter mechanism (run the
teacher, score with a verifier, keep the passing trajectories) extracts
no signal from failures. Recent advances in the training algorithm
(on-policy sampling, alternative divergences, partial
supervision;~\citealt{ICLR2024_5be69a58,luo2026agentarkdistillingmultiagentintelligence,amani2026rlreasoningadaptivelyrevealing}),
data selection~\citep{li2025naturalthoughtsselectingdistillingreasoning},
and trajectory structure~\citep{jiang2026drpdistilledreasoningpruning}
all rest on a shared assumption: the teacher can produce correct
demonstrations for the tasks that matter.

But \citet{pmlr-v267-chu25c} show that SFT tends to memorize its training
distribution rather than generalize beyond it. If the hardest scenarios
are systematically absent from that distribution, no training algorithm
or selection strategy can compensate; the ceiling on distillation quality
is set upstream, at data generation time.

\input{figures/pipeline_diagram}

In settings with execution-based verification, the standard
pipeline~\cite{chen2023fireactlanguageagentfinetuning,NEURIPS2024_61cce86d} is generate-and-filter: 
execute the teacher, score
with a verifier, and keep passing
trajectories (rejection
sampling;~\citealt{yuan2023scalingrelationshiplearningmathematical};
STaR;~\citealt{3600270.3601396};
ReST;~\citealt{gulcehre2023reinforcedselftrainingrestlanguage}). The filter is binary; a trajectory
that completed all but one of its required actions is discarded
identically to one
that failed immediately, so hard cases produce no training signal. On
$\tau$2-bench, 67\% of failed GPT-4o trajectories execute over half of
the required tool calls correctly, yet a single missed action cascades
into full failure and the entire trajectory is discarded
(Figure~\ref{fig:failure_locality}).

This asymmetry also explains why turn-level alternatives to binary
filtering are insufficient. The natural response to discarding a
near-complete trajectory is to award partial credit for the actions it
got right. But Figure~\ref{fig:failure_locality} shows tool-call
accuracy is already high in failures; rewarding correct intermediate
actions reinforces behavior the teacher already exhibits. What the
student never observes is a complete trajectory that navigates the
single decision point where the teacher fails. We confirm this
empirically in Section~\ref{sec:results}: training on the partial-credit
baseline (the filtered-only condition) lifts tool-call accuracy by 25pp
but does not improve end-to-end task completion in any model--benchmark
pair.

Likewise, sampling more, raising temperature, or running k-shot
rejection recovers stochastic failures but not strategic ones: a single
missed action at a state-dependent decision point survives any number of
resamples from the same teacher.

\input{figures/failure_locality}

We introduce \OurMethodFull{} (\OurMethod{}), which
brings prompt optimization into this pipeline but inverts its usual goal:
instead of finding one prompt that generalizes across tasks, we optimize
each scenario independently
(Figure~\ref{fig:pipeline}). Using
GEPA~\cite{agrawal2026gepareflectivepromptevolution}, an iterative prompt optimizer, we analyze
each failed execution trace and evaluation feedback, then add corrective
instructions to a per-scenario \emph{cheatsheet} appended to the
teacher's system prompt. The teacher re-executes the task from scratch,
and the process repeats until all evaluators pass. The cheatsheet is
stripped before training (Figure~\ref{fig:pipeline}); the student learns
from what the trajectory demonstrates, not from the scaffold that
produced it. The method applies to any domain with an executor and a
verifier.

This design also yields \emph{capability--voice decoupling}: because the
executor is left unchanged, recovered trajectories inherit a stronger
reflector's problem-solving without its interaction style. The resulting
distillation is \emph{voice-preserving}, letting a team upgrade task
coverage without disrupting an established product voice, latency
profile, or tone (Section~\ref{subsec:capability-voice}).

We organize our investigation around three questions: whether
per-scenario optimization reliably recovers high-quality trajectories
from teacher failures (\textbf{RQ1}); whether students trained on
recovered trajectories outperform those trained on filtered-only data,
the partial-credit baseline (\textbf{RQ2}); and whether the approach
holds up beyond benchmarks under deployment-scale evaluation across
tens of locales (\textbf{RQ3}). We address each in turn in Section~\ref{sec:results}.

We conduct our primary study on \taubench{}~\cite{barres2025tau2benchevaluatingconversationalagents}, a
stochastic multi-turn tool-calling benchmark, with additional experiments
on BFCL v4 multi-turn~\cite{pmlr-v267-patil25a}. The teacher is GPT-4o,
whose low pass rate on the official
benchmark\footnote{\url{https://llm-stats.com/models/compare/gpt-4o-2024-08-06};
GPT-4o scores 23.5\% on telecom, 45.5\% airline, 63.4\% retail.}
leaves substantial room to validate recovery. Two small
instruction-tuned models fine-tuned on recovered data improve on both
benchmarks, and the pipeline has been adopted within a
production tool-calling agent post-training system; we report all three
settings in Section~\ref{sec:results}.

%% file: figures/pipeline_diagram.tex
% Pipeline diagram — compact single-column TikZ.
% Styling matches figures/golden_trajectory_figure.tex.
% Requires in main preamble:
%   \usepackage{tikz}
%   \usepackage{amssymb}
%   \usetikzlibrary{arrows.meta, positioning, shapes.geometric, fit, calc, backgrounds}
\begin{figure}[t]
\centering
\resizebox{\columnwidth}{!}{%
\begin{tikzpicture}[
    font=\scriptsize\sffamily,
    >={Stealth[length=3pt,width=2.2pt]},
    node distance=2.5mm and 6mm,
    box/.style       = {rounded corners=1.5pt, draw=black!40, line width=0.4pt,
                        inner sep=2pt, align=center, fill=white,
                        minimum height=5.5mm, minimum width=12mm,
                        font=\scriptsize\sffamily},
    tag/.style       = {font=\scriptsize\sffamily,
                        text=black!55, inner sep=1pt},
    pipestep/.style  = {box},
    trajstep/.style  = {box, fill=blue!6,    draw=blue!60},
    optstep/.style   = {box, fill=green!8,   draw=green!55!black},
    goldstep/.style  = {box, fill=yellow!12, draw=yellow!55!black},
    outstep/.style   = {box, fill=blue!6,    draw=blue!60, line width=0.6pt},
    loopbox/.style   = {draw=black!30, dashed, rounded corners=2pt, inner sep=1.5mm},
    flow/.style      = {->, line width=0.5pt, black!85},
    rflow/.style     = {->, line width=0.4pt, black!55},
    lbl/.style       = {font=\fontsize{5.5}{6.5}\selectfont\sffamily, text=black!60,
                        inner sep=1pt},
]

% ===== Main pipeline =====
\node[pipestep]                          (tasks)  {Tasks};
\node[trajstep, right=of tasks]      (sim)    {Trajectory\\Simulation};
\node[pipestep, minimum width=14mm, right=of sim]        (scaff)  {\shortstack{Scaffold\\Removal}};
\node[goldstep, right=of scaff]      (golden) {$\bigstar$\,Golden\\Trajectories};

% ===== Per-scenario optimization loop =====
\node[optstep, below=5mm of sim, minimum width=18mm] (opt) {Per-Scenario\\Optimization};

% ===== Training (second row, under Golden Trajectories) =====
\node[outstep, below=5mm of golden] (sft) {SFT};

\draw[rflow] ([xshift=-4pt]sim.south) -- node[lbl, left, inner sep=1pt] (lblfb) {feedback} ([xshift=-4pt]opt.north);
\draw[rflow] ([xshift=4pt]opt.north)  -- node[lbl, right, inner sep=1pt] (lblcs) {cheatsheet} ([xshift=4pt]sim.south);

\begin{pgfonlayer}{background}
  % Box edges at the midpoints of the Tasks->Sim and Scaff->Golden gaps,
  % nudged out symmetrically so the opt box and feedback label are enclosed.
  \coordinate (Lx) at ($(tasks.east)!0.5!(sim.west)+(-1.6mm,0)$);
  \coordinate (Rx) at ($(scaff.east)!0.5!(golden.west)+(1.6mm,0)$);
  \draw[loopbox] ($(Lx |- sim.north)+(0,1.8mm)$)
                 rectangle ($(Rx |- opt.south)+(0,-1.8mm)$);
  \coordinate (looptop) at
      ($($(Lx |- sim.north)!0.5!(Rx |- sim.north)$)+(0,1.8mm)$);
\end{pgfonlayer}

% ===== Section tags =====
\node[tag, above=0.5mm of tasks]  {Input};
\node[tag, above=0.5mm of looptop] {\bfseries\OurMethod{}};
\node[tag, above=0.5mm of golden] {Data};

% ===== Main flow arrows =====
\draw[flow] (tasks)  -- (sim);
\draw[flow] (sim)    -- node[lbl, above, inner sep=1pt]{pass} (scaff);
\draw[flow] (scaff)  -- (golden);
\draw[flow] (golden) -- (sft);

\end{tikzpicture}%
}
\caption{Training pipeline. For each task the teacher originally failed,
a per-scenario optimization loop (green) iteratively refines a cheatsheet
from the trajectory's feedback until the teacher produces a passing
(``golden'') trajectory. The cheatsheet is stripped at scaffold removal,
so the student trains only on what the trajectory demonstrates; the
fine-tuning procedure is otherwise unchanged.}
\label{fig:pipeline}
\end{figure}
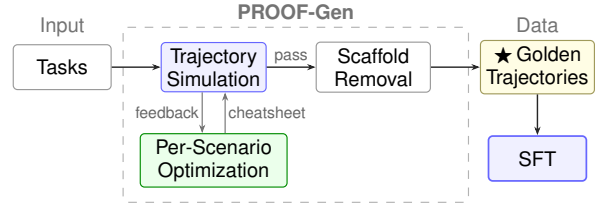

%% file: figures/failure_locality.tex
% Failure-locality figure (Figure 2) — LaTeX/TikZ version.
% Two side-by-side cascade bar panels; replaces the matplotlib
% failure_locality.pdf so fonts and layout are controlled in-document.
% Authored to fill \columnwidth with minimal whitespace; \scriptsize text
% matches the box text in figures/pipeline_diagram.tex (pipeline is scaled
% to \columnwidth at ~1.1x, so its effective box font is ~8.8pt; this figure
% stays at or below that).
% Requires in main preamble (already present):
%   \usepackage{tikz}
%   \usetikzlibrary{positioning, calc}
%
% Numbers (unchanged from the prior figure):
%   Official benchmark (473 failed sims): Tool 67%, Env 7%,  DB 6%
%   Extended telecom (2,100 failed tasks): Tool 71%, Env 3%, DB 0.1%
\begin{figure}[t]
\centering
\definecolor{flblue}{HTML}{0071E3}
\definecolor{flgray}{HTML}{86868B}
\definecolor{fldark}{HTML}{1D1D1F}
% \flpanel{x offset}{title (use \\ for line break)}{tool %}{env %}{db %}{db label}
\newcommand{\flpanel}[6]{%
  \def\flscale{0.0265}% 100% -> 2.65cm  (long bars use the column, cut empty space)
  \def\flbh{0.165}%     bar half-height (total bar thickness = 0.33cm)
  \def\flsp{0.50}%      vertical spacing between bar rows
  \node[font=\footnotesize\sffamily\color{fldark}, anchor=south, align=center]
    at ({#1+0.95},{\flbh+0.14}) {#2};
  \foreach \cat/\val/\col/\lab [count=\r from 0] in {%
      {Tool calls}/{#3}/{flblue}/{#3\%},
      {Env state}/{#4}/{flgray}/{#4\%},
      {DB state}/{#5}/{flgray}/{#6}}{%
    \pgfmathsetmacro{\yy}{-\flsp*\r}%
    \node[font=\footnotesize\sffamily\color{fldark}, anchor=east] at ({#1-0.06},\yy) {\cat};
    \fill[\col] ({#1},{\yy-\flbh}) rectangle ({#1+\val*\flscale},{\yy+\flbh});
    \node[font=\footnotesize\sffamily\bfseries\color{fldark}, anchor=west]
      at ({#1+\val*\flscale+0.07},\yy) {\lab};
  }%
}
\resizebox{\columnwidth}{!}{%
\begin{tikzpicture}
  \flpanel{0}{Official benchmark\\(473 failed sims)}{67}{7}{6}{6\%}
  \flpanel{4.3}{Extended telecom\\(2,100 failed tasks)}{71}{3}{0.1}{0.1\%}
\end{tikzpicture}%
}
\caption{The accuracy asymmetry in failed GPT-4o trajectories on $\tau$2-bench (278 tasks, 3 trials each; 473 of 834 trials fail). Most failures execute over half of tool calls correctly (high tool-call accuracy), yet a single missed action cascades into an incorrect end state (environment and database failure). Because tool-call accuracy is already high in failures, turn-level or partial-credit signals saturate without lifting end-to-end completion; what is missing is a complete trajectory through the decisive step. We confirm this in Section~\ref{sec:results}.}
\label{fig:failure_locality}
\end{figure}
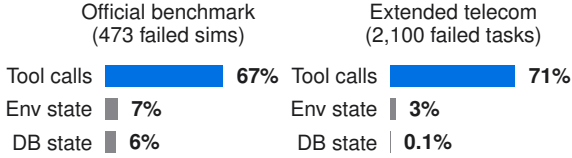

%% file: sections/method.tex
\section{Method}
\label{sec:method}

\OurMethod{} adds one step to the standard generate-and-filter pipeline:
it recovers the trajectories that filtering discards. As usual, the
teacher generates one trajectory per task and we keep the passing ones
(the \emph{filtered} set). For each \emph{failed} task, we then optimize
a disposable, per-scenario prompt until the teacher succeeds, strip that
prompt, and add the recovered trajectory to the training set
(Figure~\ref{fig:pipeline}). The student is fine-tuned on the union of
filtered and recovered trajectories.

\paragraph{Per-scenario recovery.}
A reflector model reads the failed execution trace and the verifier's
feedback (which dimensions failed, and on which assertions), then writes
a \emph{cheatsheet}: a block of natural-language guidance appended to the
teacher's system prompt for that one task. The teacher re-executes from
scratch; if it now passes every evaluator the trajectory is kept,
otherwise the reflector revises the cheatsheet and the teacher tries
again, up to $K{=}10$ iterations (each iteration sees the latest trace
and feedback, and the cheatsheet accumulates the diagnoses so far). We
use GEPA~\citep{agrawal2026gepareflectivepromptevolution} as the optimizer, though any prompt
optimizer would serve. This inverts the usual goal of prompt
optimization: rather than searching for one prompt that generalizes
across inputs, we overfit a prompt to a single scenario and discard it.
The cheatsheet is the means; the passing trajectory is the product. Its
guidance is procedural (e.g., the order of diagnostic tool calls), not
task-specific values (Figure~\ref{fig:cheatsheet-gpt51}). Scenarios still
unsolved after $K$ iterations are discarded; recovery rates are reported
in Section~\ref{sec:results}.

\paragraph{Scaffold removal.}
The cheatsheet is appended with a fixed delimiter and deleted before
training, so recovered trajectories are indistinguishable from filtered
ones and the student never sees per-scenario scaffolding. Any gain after
fine-tuning therefore reflects what the trajectories demonstrate, not the
prompt that produced them; we confirm the cheatsheets supply guidance
rather than answers in Section~\ref{subsec:datagen-quality}.

\paragraph{Configurable objective.}
Because the reflector hill-climbs on whatever the verifier reports, the
same loop can target any combination of evaluable quality dimensions
(e.g., groundedness, brevity, formatting), not only task completion; we
use this in the production setting (Section~\ref{sec:results}).

%% file: sections/experimental_setup.tex
\section{Experimental Setup}
\label{sec:experimental_setup}

We evaluate on two public benchmarks whose structure mirrors production
tool-calling post-training systems: scenarios define user requests,
execution-based evaluators serve as quality checks, and a teacher model
generates candidate trajectories. The key differences in production are
scale (thousands of scenarios with longer trajectories), multi-objective
optimization (multiple evaluators must pass simultaneously), and
continuous training (daily data ingestion and retraining); we address
these in Section~\ref{sec:results}.

\paragraph{Benchmarks.}
$\tau$2-bench~\cite{barres2025tau2benchevaluatingconversationalagents} simulates multi-turn customer
service interactions with an LLM user simulator, making evaluation
stochastic. We use the telecom domain (2{,}285 tasks, ${\sim}$50 tools)
for primary experiments: it has the largest task space and the lowest
pass rate under generate-and-filter (7.3\%), maximizing the number of
failures available for recovery (RQ1). The split is at the task level:
114 held-out base tasks form the evaluation set and the 2{,}171 non-base
tasks are used for training, with no task instance shared. Because
$\tau$2-bench tasks are parameterized from templates, train and eval may
instantiate shared templates with different parameters; we do not enforce
template-level disjointness (Appendix~\ref{app:data-composition}).
BFCL v4 multi-turn~\cite{pmlr-v267-patil25a} evaluates function-calling
across 162 functions with deterministic (pre-scripted) user turns,
spanning four categories that test distinct failure modes
(Appendix~\ref{app:bfcl-categories}). Each category contains 200 tasks
(800 total); we apply a per-category 70/30 split, giving 560 training and
240 held-out test tasks. We include BFCL to confirm that
downstream gains (RQ2) generalize beyond a single stochastic benchmark.
Table~\ref{tab:benchmarks} summarizes both.

\begin{table}[t]
\centering
\small
\begin{tabular}{lcc}
\toprule
 & $\tau$2-bench & BFCL v4 \\
\midrule
Eval tasks & 114 & 240 \\
User simulation & Stochastic & Deterministic \\
Primary metric & Pass\textasciicircum{}1 & Task accuracy \\
Secondary metric & Partial reward & Partial reward \\
\bottomrule
\end{tabular}
\caption{Benchmark comparison. $\tau$2-bench uses an LLM user simulator;
BFCL uses pre-scripted user turns.}
\label{tab:benchmarks}
\end{table}

\paragraph{Models.}
We fine-tune two small instruction-tuned models at comparable effective
scale: Gemma~4 E4B-it~\cite{googledeepmind2026gemma4e4bit} (4.5B effective
parameters) and Qwen3-4B-Instruct-2507~\cite{yang2025qwen3technicalreport} (4B
parameters). They differ in architecture and tool-call format, so
improvements on both suggest the approach is not model-specific
(Appendix~\ref{app:model-comparison}).
The teacher is GPT-4o;\footnote{\texttt{gpt-4o-2024-08-06} throughout.}
the primary reflector is GPT-5.1 for
$\tau$2-bench and GPT-5.4 for BFCL (both at high reasoning effort); each
is the strongest reflector available when that benchmark was run.

\paragraph{Data generation.}
GPT-4o generates one trajectory per task at temperature~0.
Table~\ref{tab:data-pipeline} summarizes the resulting pipeline for both
benchmarks. The teacher pass rate differs sharply (7.3\% on telecom
vs.\ 52.9\% on BFCL), and the recovery rate scales inversely with it:
per-scenario optimization recovers 93.0\% of attempted telecom failures
but 33.7\% on BFCL, where the teacher already solves the easier tasks
and only the hardest failures remain. Telecom yields 2{,}013 failures,
so we attempt recovery on a fixed random sample of 300, keeping
teacher-passing trajectories a meaningful share of the combined set
(36\%); BFCL yields only 264, small enough to attempt in full. Full
per-category counts and train/test splits are in
Appendix~\ref{app:data-composition}.

\begin{table}[t]
\centering
\small
\begin{tabular}{lrr}
\toprule
 & \textbf{\taubench{}} & \textbf{BFCL v4} \\
 & \textbf{telecom} & \textbf{multi-turn} \\
\midrule
\multicolumn{3}{l}{\textit{Generate-and-filter (teacher)}} \\
Training pool (tasks)        & 2{,}171 & 560    \\
\quad Filtered (passed)      & 158     & 296    \\
\quad Failed                 & 2{,}013 & 264    \\
Teacher pass rate            & 7.3\%   & 52.9\% \\
\addlinespace
\multicolumn{3}{l}{\textit{Per-scenario recovery}} \\
Failures sent to optimizer   & 300     & 264    \\
\quad Recovered              & 279     & 89     \\
Recovery rate                & 93.0\%  & 33.7\% \\
\addlinespace
\multicolumn{3}{l}{\textit{Combined training set (trajectories)}} \\
Total (filtered\,+\,recovered) & 437   & 385    \\
\quad From recovery          & 64\%    & 23\%   \\
\bottomrule
\end{tabular}
\caption{Data-generation pipeline for both benchmarks. The teacher is
GPT-4o (temperature~0, one trajectory per task). Telecom recovery runs
on a 300-failure sample of the 2{,}013 failures; all 264 BFCL failures
are attempted. Recovery rate = recovered / failures sent to optimizer.
The combined set merges filtered and recovered trajectories;
``from recovery'' is the recovered fraction of that set.}
\label{tab:data-pipeline}
\end{table}

\paragraph{Training.}
QLoRA~\cite{3666122.3666563} for 3 epochs on both models.
Hyperparameters are in Appendix~\ref{app:hyperparams}.

\paragraph{Evaluation.}
End-to-end task success is the primary metric. $\tau$2-bench evaluates
along three programmatic dimensions (action, environment, and database;
defined in Appendix~\ref{app:failure-analysis}), with telecom using
environment assertions as the pass/fail criterion. We run 3 trials at
temperature~0; Pass\textasciicircum{}1 is the fraction of tasks passing at least
once. At temperature~0 the agent decodes deterministically; the
3 trials average over the LLM user simulator and vLLM serving
non-determinism.
BFCL reports task accuracy and turn-level partial reward.
Sampling robustness is tested at temperature~1 on flip
tasks~\cite{zhai2026doesrlexpandcapability}.

%% file: sections/results.tex
\section{Results}
\label{sec:results}

We organize our results around three research questions:\\
\textbf{RQ1}: Does \OurMethod{} reliably recover high-quality trajectories from teacher failures?\\
\textbf{RQ2}: Do models trained on recovered trajectories outperform those trained on filtered-only data?\\
\textbf{RQ3}: Does \OurMethod{} scale to production systems that continuously do SFT retraining on large-scale datasets with multiple objectives?

\begin{figure}[t]
\centering
\includegraphics[width=\columnwidth]{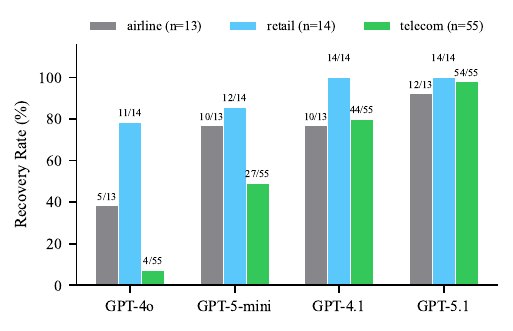}
\caption{Reflector capability dominates recovery rate (range
24.4\%--97.6\%). Per-scenario recovery by reflection model on a
cross-domain ablation set of 82 failed \taubench{} scenarios (13
airline, 14 retail, 55 telecom), holding the executor (GPT-4o) and
optimizer fixed. GPT-5.1 recovers 80/82 (97.6\%) versus 20/82 (24.4\%)
for GPT-4o self-reflection; the largest gap is on telecom (54/55 vs.\
4/55). This ablation isolates the effect of reflector capability and is
not the source of the production-scale headline rate (93\%, 279/300
telecom).}
\label{fig:gepa-recovery}
\end{figure}

\paragraph{RQ1: Recovery effectiveness.}
At production scale, per-scenario optimization recovers 93\% of sampled
telecom failures (279 of 300; GPT-4o executor, GPT-5.1 reflector;
Table~\ref{tab:frontier-comparison}); 264 (88\%) are driven by
the optimized cheatsheet and 15 by stochastic replay of the user
simulator. We report the inclusive 93\% as the headline rate and the
cheatsheet-attributable 88\% where the mechanism is the focus.
Recovery is driven by reflector capability: on a
cross-domain ablation set of 82 failures spanning all three \taubench{}
domains, it ranges from 24.4\% (GPT-4o self-reflection) to 97.6\%
(GPT-5.1; Figure~\ref{fig:gepa-recovery}). Convergence is fast: 58.7\% of
cheatsheet-driven recoveries resolve on the first iteration and 84.1\% by
the third (Figure~\ref{fig:cumulative-recovery}). Even GPT-5.1 at maximum
reasoning budget benefits from one reflection step (83.3\% $\to$ 98.3\%
on the 300 telecom tasks; Table~\ref{tab:frontier-comparison}).
Recovered trajectories constitute 64\% of the combined \taubench{}
training set; on BFCL, 23\%. The cheatsheet changes what the teacher
does, not how it communicates
(Section~\ref{subsec:capability-voice}).
Integrity audits confirm no factual leakage
(Section~\ref{subsec:datagen-quality}).
\\[2pt]
\textit{Details: Appendix~\ref{app:detailed-results} (recovery rates,
convergence, frontier comparison);
Appendix~\ref{app:analyses} (data quality audit, response complexity).}

\begin{figure*}[t]
\centering
\includegraphics[width=\textwidth]{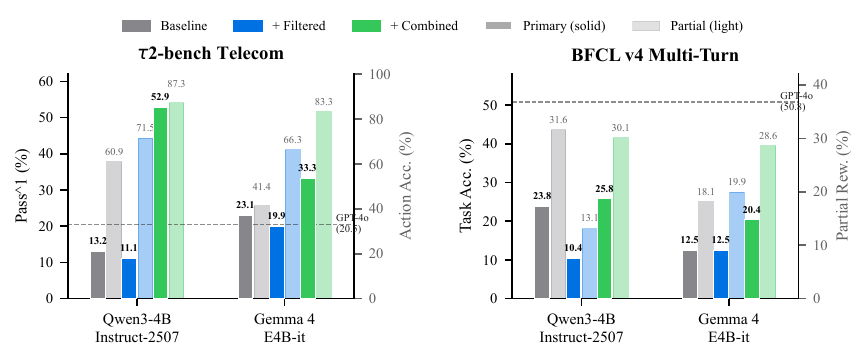}
\caption{Post-training results on \taubench{} telecom (left) and BFCL
v4 multi-turn (right). Solid bars show the primary metric (Pass\textasciicircum{}1,
task accuracy; left axis); hatched bars show partial rewards (action
accuracy, turn-level partial reward; right axis). Dashed lines mark
the GPT-4o teacher. Filtered-only training improves partial rewards
but does not improve end-to-end task completion in any of the four
model--benchmark pairs; combined training produces the only consistent
gains.}
\label{fig:main-results}
\end{figure*}

\paragraph{RQ2: Downstream impact.}
The filtered-only condition is the partial-credit baseline: it trains on
exactly the turn-level signal that is already high in failures
(Figure~\ref{fig:failure_locality}). As predicted, filtered-only
training improves partial metrics (action accuracy +25pp) but does not
improve task completion in any of the four model--benchmark pairs
(Figure~\ref{fig:main-results}); raising intermediate-action accuracy is
not sufficient when the missing signal is a complete trajectory through
the decision point. Combined training produces the
only consistent gains: Qwen3-4B-Instruct-2507~\cite{yang2025qwen3technicalreport}
improves from Pass\textasciicircum{}1=0.132 to 0.529 on \taubench{}; Gemma~4
E4B-it~\cite{googledeepmind2026gemma4e4bit} gains +7.2pp on BFCL. The
recovered trajectories cover tasks the
teacher failed without the cheatsheet (GPT-4o scores 0.205 on these
tasks); produced by GPT-4o under a stronger reflector, they let the
student inherit problem-solving the base teacher does not exhibit on its
own. Gains persist
under stochastic sampling at temperature~1: on flip tasks (post-hoc
selected where fine-tuning helped at temp=0; $n{=}46$ Qwen, $n{=}21$
Gemma), the combined model averages 54.1\% (Qwen) and 39.0\% (Gemma)
per-trial success, versus near-zero baselines
(Figure~\ref{fig:sampling-robustness}).

\paragraph{Capability vs.\ voice: why the executor matters.}
\label{subsec:capability-voice}
Per-scenario optimization separates two factors that direct distillation
conflates: the reflector supplies task-solving \emph{capability}, while
the executor supplies interaction \emph{voice}. On 259 matched telecom
tasks, the cheatsheet changes which tools GPT-4o calls but not how it
communicates: GPT-4o keeps its terse, incremental turn structure (one
diagnostic at a time), whereas GPT-5.1 front-loads a single numbered
multi-step protocol 3.8$\times$ longer at identical tool-call counts
(Appendix~\ref{subsec:response-complexity},
\ref{app:trajectory-patterns}). \OurMethod{} therefore yields
\emph{voice-preserving distillation}: a team can adopt a stronger
reflector's problem-solving without inheriting its voice or latency
profile.

\paragraph{RQ3: Production scaling.}
To test whether \OurMethod{} transfers beyond the controlled \taubench{}
and BFCL benchmarks, we deployed it in a production tool-calling agent
post-training system. Evaluation is end-to-end: thousands of held-out
scenarios are run in simulation and scored by deterministic verifiers
together with LLM judges, across goal completion and five
response-quality metrics (entity formatting, brevity, groundedness,
tool-calling accuracy, style and tone). Absolute values are withheld per
deployment policy, so we report percentage-point (pp) gains over the
respective baselines; as the production evaluation is a single
large-scale run, we treat consistency across metrics and locales as the
robustness signal. Per-scenario
optimization improved generated training-trajectory quality by +6.3pp in
goal completion and +2.5 to +8.0pp across the five response-quality
metrics. These gains transferred to the downstream on-device model,
evaluated without any cheatsheet augmentation: +1.5pp goal completion and
+1.7 to +5.0pp across the same metrics, with no regression on any
dimension. The effect holds across tens of locales (positive transfer in
every evaluated locale; non-English average +1.48pp goal completion,
range +0.15 to +3.38pp), indicating the corrective knowledge is largely
language-agnostic. We present this as a deployment case study
corroborating the controlled, reproducible benchmark results.

\paragraph{Efficiency and orchestration.}
Per-scenario optimization is fully parallel: scenarios are optimized
independently with no shared state, so the recovery stage fans out across
workers. It is also cheap in frontier calls: the expensive reflector only
diagnoses and revises the cheatsheet while the cheaper executor handles
the rollout, far fewer frontier calls than using a frontier model as the
executor at every turn (Appendix~\ref{subsec:frontier-comparison}).
Implementation is standard: an off-the-shelf prompt optimizer (GEPA via
DSPy) wrapped in any job scheduler (e.g., Ray, Airflow).

%% file: sections/related_work.tex
\section{Related Work}
\label{sec:related_work}

\OurMethod{} addresses the coverage gap that generate-and-filter leaves
in every training cycle; we situate it against three lines of work.

\paragraph{SFT for tool-calling agents.}
Training small models to use tools via SFT on teacher-generated
trajectories is well established (ToolLLM~\cite{ICLR2024_28e50ee5},
FireAct~\cite{chen2023fireactlanguageagentfinetuning}, APIGen~\cite{NEURIPS2024_61cce86d},
xLAM~\cite{zhang-etal-2025-xlam}, and Magnet~\cite{yin-etal-2025-magnet}). All treat
data generation as uniform sampling from the teacher, keeping whatever
passes and discarding the rest; our contribution is recovering the
failures.

\paragraph{Trajectory recovery and self-improvement.}
Methods that reuse failures (STaR~\cite{3600270.3601396},
ReST$^{\text{EM}}$~\cite{singh2024beyond},
V-STaR~\cite{hosseini2024vstar}, ETO~\cite{song-etal-2024-trial},
Agent-R~\cite{yuan2025agentrtraininglanguagemodel}, and Reflexion~\cite{3666122.3666499})
share a structural assumption: a successful trajectory already exists for
each failure, supplied by an expert, a stronger model, or a passing
rollout to pair against (offline preference learning in V-STaR and
ETO). \OurMethod{} targets the upstream problem: it \emph{creates} a
successful trajectory where no expert solves the task, modifying only the
teacher's prompt (no weight updates, tree expansion, or failure examples).
The two are complementary: recovered trajectories
can feed any preference method that needs paired data.

\paragraph{Prompt optimization as a data-generation instrument.}
Automatic prompt optimization typically searches for one prompt that
generalizes across inputs at inference
time~\cite{pryzant-etal-2023-automatic,ICLR2024_3339f19c,opsahl-ong-etal-2024-optimizing,ICLR2024_f1cf02ce,ICLR2024_82eec786}.
ACE~\cite{zhang2026agenticcontextengineeringevolving} is the closest analog, accumulating task-specific
``playbooks'' from execution feedback like our cheatsheets, but deploys
them as persistent inference-time context. We instead treat the prompt as
a disposable instrument for exhausting a single scenario's solution space
and discard it before training, retaining only the passing trajectory; to
our knowledge, using automatic prompt optimization explicitly as a
data-generation engine has not been previously studied.

%% file: sections/conclusion.tex
\section{Conclusion}
\label{sec:conclusion}

Generate-and-filter distillation discards the teacher's hardest
failures, leaving a coverage gap no downstream training algorithm
can close. \OurMethod{} treats them as recoverable: per-scenario
prompt optimization steers the teacher to a passing trajectory, and the
scaffold is stripped so the student learns from clean demonstrations.
It recovers 93\% of sampled telecom failures, lifts
downstream task completion for two small instruction-tuned students on
\taubench{} and BFCL (the only condition to do so in all four pairs), and
improves a production agent post-training system with positive transfer
in every locale.

%% file: sections/limitations.tex
\section*{Limitations}
\label{sec:limitations}

\paragraph{Reflector generalization.}
We characterize the recovery-vs-capability relationship across four
models from the same provider. Extending this study to other providers
and open-source models remains future work.

\paragraph{Teacher model scope.}
We use GPT-4o as the sole teacher. Section~\ref{subsec:response-complexity}
shows that the teacher's response style shapes the training data
independently of task coverage. An open direction is self-distillation:
using the student model itself as the teacher, recovering its failures
via per-scenario optimization, and fine-tuning on its own corrected
trajectories. Whether this preserves native response behavior while
expanding task coverage, compared to cross-model distillation, is an
open research question.

\paragraph{Instruction-following assumption.}
The load-bearing assumption of the method is that the executor can
follow long, structured cheatsheets: the cheatsheets that drive recovery
reach 7--9K characters with explicit step numbering, tool-call patterns,
and conditional branches (Appendix~\ref{app:cheatsheet}). A capable
frontier executor (GPT-4o) satisfies this, but the assumption becomes
binding for the self-distillation extension above, where a 4B student
must execute the same frontier-written cheatsheets; its in-context
instruction-following capacity sets an upper bound on how many failures
self-distillation can recover. We have not measured this fraction
directly; running frontier-written cheatsheets through the student
executor would quantify the assumption and is a prerequisite for that
extension.

\paragraph{Benchmark constraints.}
Both \taubench{} and BFCL are designed as evaluation benchmarks; we
repurpose their executors and verifiers into a data generation pipeline
with separate train/eval splits. This limits data scale: only
\taubench{} telecom has sufficient task volume (2{,}285 tasks) for our
experiments, while airline and retail are too small for end-to-end
validation. On BFCL, tasks span 8 API domains and 4 evaluation
categories, but each category contains only 200 tasks (800 total): the
model must acquire multi-domain, multi-category tool-calling
capabilities from very few examples per category. Per-category results at $n{=}60$ (test split, 70/30)
reflect this: \texttt{miss\_func} (which tests a capability the baseline
largely lacks) shows consistent gains for both models, while other
categories show mixed effects that are directional rather than
definitive.

%% file: sections/ethics.tex
\section*{Ethics Statement}

To the best of our knowledge, all results reported in this paper are
accurate. The two public benchmarks we use, \taubench{} and BFCL, are
openly available and cited accordingly, as are the teacher, reflector,
and student models.

The production study in Section~\ref{sec:results} is evaluated on an
internal held-out set of synthetic scenarios that simulate production use
cases. These scenarios are authored and human-reviewed for evaluation
purposes; they are not sampled from real user traffic and contain no
personally identifiable information. No human subjects were involved, and
no private user data was used at any stage of data generation, training,
or evaluation. Absolute performance values for the production system are
withheld per deployment policy, so we report percentage-point gains over
the respective baselines instead.

%% file: sections/acknowledgments.tex
\section*{Acknowledgments}

We thank Frankie Liu for contributions to the exploratory experiments in
this study and for extensive work on the internal product. We are
grateful to Sylvia Xu for detailed feedback on multiple drafts of this
paper. We also thank Anatoly Adamov, Alex Braunstein, and Amar
Subramanya for their continued support of both the research and its path
into production.

%% file: sections/appendix.tex
\appendix

\section{Detailed Benchmark Results}
\label{app:detailed-results}

This appendix expands the per-question summary in
Section~\ref{sec:results} with the full breakdowns it refers to: the
failure analysis the method targets (\ref{app:failure-analysis}),
recovery by reflector and convergence (\ref{subsec:datagen-results}), the
frontier-model comparison (\ref{subsec:frontier-comparison}), and the
post-training results (\ref{subsec:post-training}).

\subsection{Failure Analysis}
\label{app:failure-analysis}

We characterize the failures that per-scenario optimization targets.
On the $\tau$2-bench base split (50 airline, 114 retail, 114 telecom;
3 trials each, 834 total), 473 trials fail. Of these, 67\% execute over
half of the required tool calls correctly, yet only 7\% reach the
correct environment state and 6\% produce the correct database state
(Figure~\ref{fig:failure_locality}). Tool call accuracy ranges from 51\%
in airline to 71\% in telecom.

We confirmed this at larger scale: on 2{,}285 telecom tasks (1 trial
each), 2{,}100 fail. Of these, 71\% get over half of tool calls right,
while environment and database correctness drop to 3\% and 0.1\%.

\begin{figure}[t]
\centering
\includegraphics[width=\columnwidth]{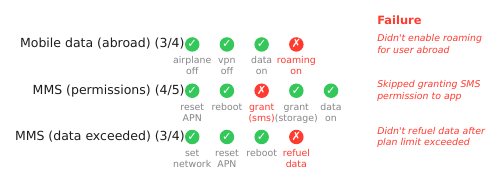}
\caption{Three failed GPT-4o trajectories from $\tau$2-bench telecom. Green circles indicate tool calls that matched the ground truth; red circles indicate missed calls. Each trajectory completes most required actions but fails at one decision point, causing the entire trajectory to be discarded by the filter.}
\label{fig:failure_examples}
\end{figure}

These failures share a common structure: the teacher completes routine
operations correctly, then misses a single action at a decision point
that depends on state accumulated earlier in the trajectory. Because
$\tau$2-bench evaluates end-to-end correctness, one missed tool call
cascades into incorrect environment and database state, and the entire
trajectory is discarded. Figure~\ref{fig:failure_examples} shows three
concrete examples: a skipped roaming toggle for a customer abroad, a
missing SMS permission grant for an MMS issue, and an overlooked data
refuel after the plan limit was exceeded. In each case, 3 or 4 out of
4--5 required actions were correct.

\subsection{Data Generation}
\label{subsec:datagen-results}

\paragraph{Reflection model capability drives recovery rate.}

Recovery effectiveness depends on the reflection model that analyzes failed trajectories and generates cheatsheets. We ablated four reflection models on 82 scenarios that failed all 3 trials under GPT-4o across three \taubench{} domains (Figure~\ref{fig:gepa-recovery}). All other variables are held constant: the agent model (GPT-4o, temp=0), user simulator (GPT-4.1, temp=0), and optimizer configuration (max 10 iterations, stop at reward=1.0).

Reflection model capability is the dominant factor. GPT-5.1 recovers 97.6\% of failed scenarios (80/82) versus 24.4\% for GPT-4o (20/82, self-reflection). The gap is largest on telecom, the most complex domain (4/55 vs.\ 54/55). Recovery counts include both cheatsheet-driven improvements and cases where the teacher succeeded on stochastic replay without a cheatsheet; Appendix~\ref{app:recovery-breakdown} reports the breakdown.

\paragraph{Convergence behavior.}

\begin{figure}[t]
\centering
\includegraphics[width=\columnwidth]{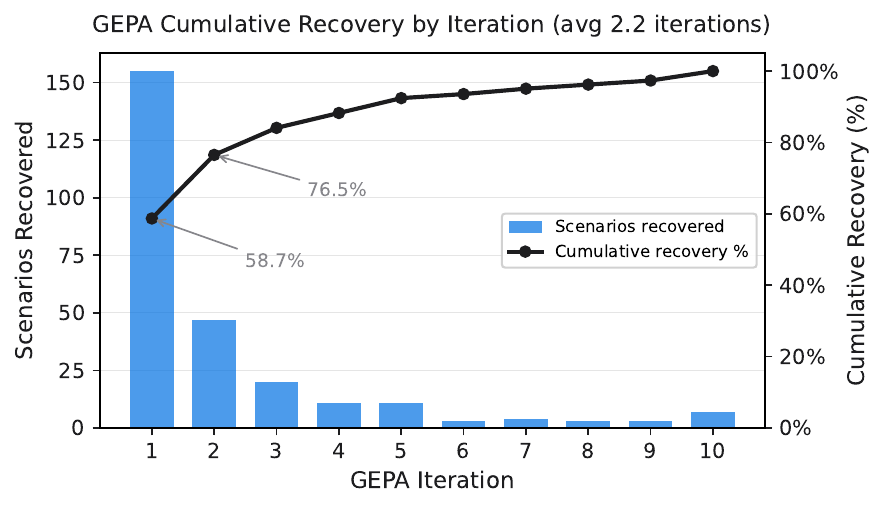}
\caption{Cumulative recovery by optimizer iteration on 300 failed telecom tasks (GPT-4o agent, GPT-5.1 reflector). 264 of 279 total recoveries are cheatsheet-driven; the remaining 15 are stochastic replays. Most solvable scenarios resolve within 2--3 iterations.}
\label{fig:cumulative-recovery}
\end{figure}

On the 300-task telecom production run with GPT-5.1 as reflector, convergence is bimodal (Figure~\ref{fig:cumulative-recovery}). Of the 300 attempted failures, 279 are recovered (264 driven by a cheatsheet, 15 by stochastic replay without one) and 21 are never recovered. Of the cheatsheet-driven recoveries, 58.7\% resolve on the first iteration, 76.5\% by the second, and 84.1\% by the third. Scenarios that remain unsolved after 3 iterations rarely recover in later iterations. This suggests that the mechanism is cheatsheet \emph{quality} (whether the reflection correctly diagnoses the root cause) rather than brute-force search over prompt variations.

\paragraph{Data composition.}

The composition of the combined training data differs sharply between the
two benchmarks. On \taubench{}, recovered trajectories constitute 64\%
of the combined set, so training signal comes predominantly from tasks
the teacher originally failed. On BFCL, recovered trajectories
constitute only 23\%; the dataset remains composed primarily of tasks
the teacher already solves.
Appendix~\ref{app:data-composition} reports the full trajectory counts
and example volumes per benchmark and category.

\subsection{Can Frontier Models Replace Per-Scenario Optimization?}
\label{subsec:frontier-comparison}

The capable model used as the reflector (GPT-5.1) could also generate
trajectories directly, without the per-scenario optimization loop. We
compare four generation strategies on 300 telecom tasks that GPT-4o
failed in the initial generation pass (Table~\ref{tab:frontier-comparison}).

\begin{table}[t]
\centering
\small
\resizebox{\columnwidth}{!}{%
\begin{tabular}{llcc}
\toprule
\textbf{Agent} & \textbf{Reflection} & \textbf{Recovered} & \textbf{Rate} \\
\midrule
GPT-5.1 (no reasoning) & none & 44/300 & 14.7\% \\
GPT-5.1 (high)    & none & 250/300 & 83.3\% \\
GPT-4o            & GPT-5.1 (high), 10 iter & 279/300 & 93.0\% \\
GPT-5.1 (high)    & GPT-5.1 (high), 1 iter  & 295/300 & 98.3\% \\
\bottomrule
\end{tabular}%
}
\caption{Recovery rate on 300 failed telecom tasks (selected from GPT-4o
failures by design). GPT-5.1 defaults to reasoning\_effort=none (no
reasoning); ``high'' = reasoning\_effort=high. Per-scenario
optimization with a weaker agent (GPT-4o) and capable reflector recovers
93.0\% using the expensive model only for reflection (avg 2.2 calls per
scenario vs.\ ${\sim}$11.5 execution turns per task for direct
generation).}
\label{tab:frontier-comparison}
\end{table}

\paragraph{GPT-5.1 without reasoning: 14.7\% recovery.}
GPT-5.1 without reasoning solves 44 of 300 tasks (14.7\%). Upgrading
the model alone (without enabling reasoning) helps over GPT-4o, but
leaves 85\% of failures unresolved.

\paragraph{GPT-5.1 at high reasoning: 83.3\% recovery.}
GPT-5.1 at high reasoning effort solves 250 (83.3\%). This approaches
the recovery rate of per-scenario optimization with a weaker agent
(GPT-4o agent + GPT-5.1 reflector: 279/300, 93.0\%), but requires the
most expensive inference setting on every turn of every task.

\paragraph{Adding one reflection step: 98.3\% recovery.}
Adding a single reflection step to GPT-5.1 high pushes recovery
to 295/300 (98.3\%). The strongest model we tested still benefits from
per-scenario cheatsheet guidance, producing 45 additional recoveries
beyond what high-reasoning direct generation achieves alone.

\paragraph{Model usage pattern.}
In the GPT-4o executor configuration (row 3), the capable model is used
only for reflection (2.2 calls per scenario on average; 58.7\% resolve
on the first cheatsheet), while the weaker model handles multi-turn
execution. Frontier direct execution at full reasoning budget (row 2)
requires the capable model on every turn of the task.
The two strategies also produce stylistically different trajectories;
Section~\ref{subsec:response-complexity} examines how the executing
model's behavior shapes training data.

\subsection{Post-Training Results}
\label{subsec:post-training}

Figure~\ref{fig:main-results} summarizes the main results across both
models and benchmarks. Each model is evaluated under three conditions:
no fine-tuning (baseline), QLoRA on filtered-only data, and QLoRA on
the combined dataset (filtered plus recovered trajectories).

\paragraph{\taubench{}.}
On \taubench{} telecom (Figure~\ref{fig:main-results}, left),
Qwen3-4B-Instruct-2507 improves from a baseline of 0.132 to Pass\textasciicircum{}1 =
0.529 with combined training.\footnote{We verified training stability by retraining from scratch and re-evaluating multiple times. Across independent training runs and evaluations, Qwen Pass\textasciicircum{}1 ranged from 0.518 to 0.538, confirming that the reported gains are not artifacts of a favorable random seed. See Appendix~\ref{app:training-stability} for details.} For context, the GPT-4o teacher that
generated the training data scores 0.205 on the same 114-task evaluation
set under the same protocol (3 trials, temp=0).
Gemma~4 E4B-it presents a harder test: its baseline (0.231) is already
on par with the GPT-4o teacher. The method still improves it to 0.333:
although GPT-4o executes the trajectories, the recovery process is
guided by GPT-5.1 reflection, so the training signal encodes capability
beyond what the executing teacher produces alone.
Action accuracy follows the same pattern: the combined model reaches
87.3\% (Qwen) and 83.3\% (Gemma), up from 60.9\% and 41.4\%
respectively.
Section~\ref{subsec:sampling-robustness} tests whether these
improvements hold under stochastic sampling at temperature~1.

\paragraph{BFCL v4 multi-turn.}
On BFCL v4 multi-turn~\citep{pmlr-v267-patil25a}, Gemma~4 E4B-it gains +7.2
percentage points overall, while Qwen3-4B-Instruct-2507 gains +1.9pp.
Filtered-only training improves
partial rewards but does not improve end-to-end task completion over
the baseline in any of the four model--benchmark pairs. Turn-level partial reward confirms the
pattern: Gemma improves from 0.181 to 0.286 (+58\%); Qwen's partial
reward is flat (0.316 to 0.301).%
\footnote{We ran three independent evaluations per condition to assess
serving non-determinism. Overall accuracy is stable across runs (spread
$\leq$1.6pp); per-category results (60 tasks each) vary by up to 3.3pp
from this source alone, which should be considered when interpreting
category-level comparisons.}
Section~\ref{subsec:finetuning-analysis} examines why.
The gain concentrates in the \texttt{miss\_func} category (recognizing
unavailable functions), the only category where both models improve
consistently; per-category results are in
Appendix~\ref{app:bfcl-categories}.

\FloatBarrier
\subsection{Sampling Robustness}
\label{subsec:sampling-robustness}

The improvements reported above use greedy decoding (temp=0). To test
whether training gains reflect genuine capability expansion rather than
improved greedy-path reliability, we follow the framework of
\citet{zhai2026doesrlexpandcapability}. We identify ``flip tasks'' (tasks that
changed from 0/3 passes at baseline to $\geq$2/3 after fine-tuning) and
re-evaluate both models on these tasks at temperature~1.0 with 10 trials
per task. Flip tasks are post-hoc selected on temp=0 gains, so this
analysis measures robustness on the subset where fine-tuning helped, not
overall capability; the subsets are small (Qwen $n{=}46$, Gemma
$n{=}21$).

\begin{figure}[t]
\centering
\includegraphics[width=\columnwidth]{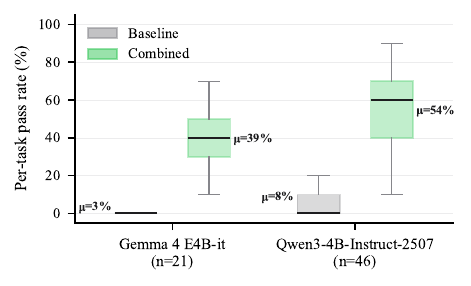}
\caption{Distribution of per-task pass rates at temperature~1 (10 trials)
on \taubench{} flip tasks. Boxes show quartiles; black lines mark means.
Baseline rates cluster near zero; combined rates are broadly distributed,
confirming that training gains persist under stochastic
sampling.}
\label{fig:sampling-robustness}
\end{figure}

Figure~\ref{fig:sampling-robustness} shows per-task pass rates on the
flip tasks. On these tasks the combined model sustains substantial
per-trial success under stochastic sampling: Qwen averages 54.1\% (46
tasks) and Gemma 39.0\% (21 tasks), versus near-zero baselines (8.3\%
and 2.9\% respectively). The pattern is consistent across individual
tasks: the combined model produces broadly distributed success rates
while the baseline is near zero almost everywhere.

\section{Additional Benchmark Analyses}
\label{app:analyses}

\subsection{Data Quality}
\label{subsec:datagen-quality}

\paragraph{Cheatsheet integrity: the scaffold does not create shortcuts.}

A natural concern is whether cheatsheets introduce unrealistic shortcuts into the generated trajectories. Because the cheatsheet is stripped before SFT, the student model never sees it. However, if the cheatsheet causes the teacher to bypass discovery steps (e.g., using an entity ID without first looking it up), the resulting trajectory would teach the student to hallucinate values rather than discover them through tool calls.

We audited 90 cheatsheet-guided trajectories: 60 from \taubench{} (15 per reflection model across all four reflectors) and 30 from BFCL (using the production reflector, GPT-5.4), using two independent LLM judges (Claude Opus 4.6 and Claude Sonnet 4.6). Each judge evaluated the full, untruncated trajectory against three checks: (1)~no hardcoded values (all tool call arguments sourced from the user's request, prior tool returns, or schema defaults); (2)~no magic numbers (computed values derived from prior lookups, not hardcoded); and (3)~no future information (each tool call uses only information available at the time it was called). Table~\ref{tab:cheatsheet-integrity} reports the results.

\begin{table}[t]
\centering
\small
\begin{tabular}{lcc}
\toprule
\textbf{Reflection Model} & \textbf{Opus 4.6} & \textbf{Sonnet 4.6} \\
\midrule
\multicolumn{3}{l}{\textit{\taubench{} (60 samples, 15 per teacher)}} \\
\quad GPT-4o      & 15/15 & 14/15 \\
\quad GPT-5-mini  & 15/15 & 15/15 \\
\quad GPT-4.1     & 15/15 & 15/15 \\
\quad GPT-5.1     & 15/15 & 15/15 \\
\midrule
\multicolumn{3}{l}{\textit{BFCL v4 (30 samples, GPT-5.4 teacher)}} \\
\quad All categories & 30/30 & 29/30 \\
\midrule
\textbf{Total}    & \textbf{90/90} & \textbf{88/90} \\
\bottomrule
\end{tabular}
\caption{Cheatsheet integrity audit. Each cell shows the number of trajectories rated \textsc{Transferable} (all discovery steps preserved). Sonnet's two flags are false positives on manual inspection (Appendix~\ref{app:integrity}).}
\label{tab:cheatsheet-integrity}
\end{table}

Both judges agree that all 90 trajectories preserve value discovery through tool calls (97.8\% inter-judge agreement). Two edge cases received split verdicts; manual inspection confirmed no actual leakage (details in Appendix~\ref{app:integrity}).

Qualitatively, cheatsheets teach \emph{process} (troubleshooting methodology, tool ordering, and error recovery patterns) rather than \emph{answers}. Stronger reflection models produce longer, more structured cheatsheets that describe procedures without referencing scenario-specific values, while GPT-4o's shorter cheatsheets occasionally include task-specific hints alongside discovery instructions. While this style difference does not compromise trajectory integrity (discovery instructions are present across all reflection models), it may contribute to the recovery rate gap observed in Section~\ref{subsec:datagen-results}: more structured cheatsheets appear to guide the teacher more reliably past failure points.

\subsection{Fine-Tuning Analysis}
\label{subsec:finetuning-analysis}

\paragraph{Reward breakdown.}

\taubench{} evaluates task success using a configurable set of criteria:
database checks, environment/status assertions, action matching,
communication-information checks, and natural-language assertions. In
telecom, task success is determined primarily by programmatic environment
assertions, with action matching included for transfer-style tasks;
communication checks and NL assertions are not used. This makes telecom
evaluation largely programmatic and avoids variance from LLM-judged
criteria. Table~\ref{tab:reward-breakdown} reports action, environment,
and database metrics as diagnostic breakdowns for the telecom domain.

\begin{table}[t]
\centering
\small
\begin{tabular}{lccc}
\toprule
\textbf{Experiment} & \textbf{Action} & \textbf{Env} & \textbf{DB} \\
\midrule
Baseline (no SFT)  & 41.4\% & 23.5\% & 9.3\% \\
+ QLoRA Filtered   & 66.3\% & 22.6\% & 10.9\% \\
+ QLoRA Combined    & \textbf{83.3\%} & \textbf{54.5\%} & \textbf{15.6\%} \\
\bottomrule
\end{tabular}
\caption{Reward breakdown on \taubench{} telecom (base-114 held-out tasks, 3 trials, temp=0). Filtered-only training improves action accuracy but not environment or database accuracy. Combined training improves all three, with the largest gain in environment assertions.}
\label{tab:reward-breakdown}
\end{table}

Filtered-only training improves action accuracy by +25pp but leaves environment assertions and database accuracy flat. Combined training closes this gap: environment assertions improve by +31pp and database accuracy by +6pp. The recovered trajectories expand task coverage to scenarios the teacher originally failed, adding demonstrations of tasks that require multi-step state management beyond what the teacher handles on the first attempt. Appendix~\ref{app:bfcl-categories} reports a parallel category-level breakdown for BFCL.

\subsection{Teacher Response Complexity}
\label{subsec:response-complexity}

GPT-5.1 at high reasoning matches or exceeds GPT-4o's recovery rate (Table~\ref{tab:frontier-comparison}), but the resulting trajectories differ in structure. A natural question is why not use GPT-5.1 directly as the teacher, bypassing the recovery loop entirely. We investigate by comparing GPT-4o and GPT-5.1 trajectories on matched tasks.

\paragraph{Controlled comparison on matched tasks.}

\begin{figure}[t]
\centering
\includegraphics[width=\columnwidth]{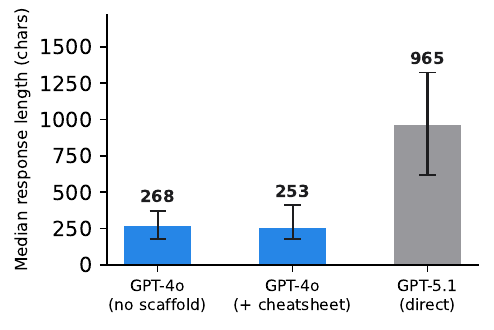}
\caption{Median text response length on matched tasks. GPT-4o without scaffold (n=15 tasks) and GPT-4o with cheatsheet (n=259) produce comparable response lengths, while GPT-5.1 responses are 3.8$\times$ longer. Whiskers show IQR.}
\label{fig:response-complexity}
\end{figure}

From the 300-task recovery set, 15 tasks pass on GPT-4o replay alone (no cheatsheet; stochastic user variation), 264 require cheatsheet guidance, and 21 are never recovered. GPT-5.1 (high reasoning, no scaffold) passes 295 of these tasks directly. We compare text responses on the 259 tasks both models solve; because task identity is held constant, differences in response characteristics reflect model behavior rather than task difficulty. Figure~\ref{fig:response-complexity} and Table~\ref{tab:response-complexity} report text response statistics on this matched set, with the 15 replay-pass tasks as a separate condition in the figure.

\begin{table}[t]
\centering
\small
\begin{tabular}{lrr}
\toprule
\textbf{Metric} & \textbf{GPT-4o} & \textbf{GPT-5.1 (high)} \\
\midrule
Trajectories & 259 & 259 \\
Text responses & 3{,}180 & 2{,}863 \\
Avg.\ length (chars) & 308 & 987 \\
Median length (chars) & 253 & 965 \\
Tool calls / trajectory & 5.2 & 5.2 \\
\bottomrule
\end{tabular}
\caption{Text response characteristics on 259 matched telecom tasks.
GPT-4o uses cheatsheet guidance; GPT-5.1 uses direct high reasoning.
Tool call counts are comparable; response length differs by
3.8$\times$.}
\label{tab:response-complexity}
\end{table}

GPT-5.1 text responses are 3.8$\times$ longer at the median (965 vs.\ 253 characters), despite comparable tool call counts (5.2 calls per trajectory). GPT-4o produces more text turns per trajectory (12.3 vs.\ 11.1), each substantially shorter: it confirms the completed action in one sentence, suggests one next step, and waits for the user to report back. GPT-5.1 front-loads a numbered multi-step protocol into a single turn, with specific device instructions and a conditional branching plan. Both approaches resolve the task; the verifier checks final state and required actions, not response phrasing. Appendix~\ref{app:trajectory-patterns} shows both models on the same scenario at a representative 3.7$\times$ ratio.

\paragraph{Scaffold effect on response style.}

Because per-scenario optimization uses GPT-4o as the executing agent, its trajectories preserve GPT-4o's incremental communication pattern even on tasks that required GPT-5.1's reflection to recover. Figure~\ref{fig:response-complexity} and the per-tool breakdown in Appendix~\ref{app:trajectory-patterns} show that the cheatsheet changes what tools GPT-4o calls but does not alter how it communicates results: GPT-4o replay passes (no cheatsheet, n=15) show a median of 268 characters, comparable to GPT-4o with cheatsheet at 253 (0.94$\times$). Neither condition approaches GPT-5.1's 965-character median.

This separates two concerns in data generation: task coverage (determined by the reflector model) and response style (determined by the executing agent), which can in principle be varied independently.

\section{Cheatsheet Integrity Audit}
\label{app:integrity}

\subsection{Judge Rubric}

Each of the 90 trajectories was evaluated independently by Claude Opus 4.6 (\texttt{claude-opus-4-6}) and Claude Sonnet 4.6 (\texttt{claude-sonnet-4-6}) using the prompt in Figure~\ref{fig:judge-rubric}. The judge receives the full, untruncated task definition, cheatsheet, and trajectory as input. The three checks are summarized below; the exact prompt wording is reproduced for full reproducibility.

\paragraph{Check 1: No hardcoded values.} For every value used in a tool call argument (entity IDs, file paths, state values, amounts), was it either (a)~provided in the user's request, (b)~returned by a prior tool call, or (c)~a reasonable default from the tool schema? Answer ``no'' only if a value could not have been obtained without reading the cheatsheet.

\paragraph{Check 2: No magic numbers.} If the trajectory uses a computed value, does it include the steps to compute or look up that value? Answer ``no'' if the trajectory uses a hardcoded number that only makes sense given the cheatsheet.

\paragraph{Check 3: No future information.} Does each tool call use information available at the time it was called? The question is whether information flows correctly, not whether the ordering is optimal or whether calls are sequential versus parallel.

\paragraph{Verdict.} \textsc{Transferable} or \textsc{Has\_Shortcut}. Parallel tool calls, action ordering preferences, missing user confirmations, and policy violations are explicitly excluded from consideration; these are execution style, not discovery bypass.

\begin{figure*}[t]
\centering
\begin{lrbox}{\judgepromptbox}
\begin{BVerbatim}
You are an expert reviewer analyzing whether a tool-calling trajectory teaches
transferable behavior, or whether it contains shortcuts that would break on a new task.

Context: A teacher model (GPT-4o) generated this trajectory with help from a
"cheatsheet" -- a natural-language hint appended to the system prompt. The cheatsheet
is stripped before SFT. The student never sees the cheatsheet.

The cheatsheet being prescriptive is fine and expected. The only concern is: does the
resulting trajectory teach a tool-calling pattern that would work on a new, similar task?

What is NOT a shortcut (do NOT flag these):
- Parallel/batched tool calls: execution style, not discovery bypass.
- Action ordering preferences: diagnostic choice, not a shortcut.
- Missing user confirmation: politeness issue, not discovery bypass.
- Policy violations: format issue, not transferability issue.

Only flag as HAS_SHORTCUT if a value used in a tool call argument was NOT discoverable
from the user's request or prior tool returns.

Inputs: [task_definition] [cheatsheet] [trajectory]

Checks (yes/no + rationale):
  1. No hardcoded values: all argument values sourced from user request, prior tool
     returns, or schema defaults? "no" ONLY if a value could not have been obtained
     without reading the cheatsheet.
  2. No magic numbers: computed values justified by prior steps? "no" if the trajectory
     uses a number that only makes sense if you read the cheatsheet.
  3. No future information: each tool call uses information available at call time?
     "no" ONLY if a tool call uses a value not yet returned by any prior call.

Verdict: TRANSFERABLE or HAS_SHORTCUT.
\end{BVerbatim}
\end{lrbox}
\resizebox{\textwidth}{!}{\usebox{\judgepromptbox}}
\caption{Exact prompt template given to both LLM judges. Placeholders \texttt{[task\_definition]}, \texttt{[cheatsheet]}, and \texttt{[trajectory]} are filled with the full, untruncated text for each sample.}
\label{fig:judge-rubric}
\end{figure*}

\subsection{Sonnet False Positive Analysis}

Sonnet flagged 2 of 90 samples as \textsc{Has\_Shortcut}. Both are false positives on manual inspection.

\paragraph{Sample 11 (\taubench{} telecom, GPT-4o teacher).} After receiving line IDs [L1001, L1002, L1003] from the customer profile, the agent fetched only L1002 directly. Sonnet flagged this as skipping the phone number matching step. However, the customer's phone number (555-123-2002) shares a suffix pattern with L1002, making this a plausible heuristic. The tool return confirms the match (\texttt{phone\_number: 555-123-2002}). The agent does not proceed without verification; it receives and acts on the confirmed match.

\paragraph{Sample 77 (BFCL miss\_func, GPT-5.4 teacher).} Sonnet flagged the use of Twitter credentials (\texttt{tech\_guru}, \texttt{securePass123}) as cheatsheet-sourced. The user's message appears truncated in the trajectory representation at ``My user name...'' However, BFCL provides fixed, deterministic user messages, and the full message includes both username and password. The truncation is an artifact of the trajectory display format, not the actual conversation the agent processed.

\subsection{Cheatsheet Examples by Reflection Model}

Representative cheatsheets from four reflection models on \taubench{} telecom scenarios illustrate the relationship between teacher capability and cheatsheet style.

\paragraph{GPT-4o (1.8K chars).} Short, mixes task-specific hints with discovery instructions. Example: ``Identify relevant line: Use the correct line ID associated with the phone number, in this case, L1002.'' The discovery instruction is present but the specific value is named.

\paragraph{GPT-5-mini (9.8K chars).} Verbose, focuses on behavioral patterns. Identifies wrong \emph{behaviors} (e.g., ``the assistant did not correctly refuel data'') and prescribes corrective actions.

\paragraph{GPT-4.1 (3.7K chars).} Mid-length, structured. Separates failure diagnosis from step-by-step guidance. Uses parameterized instructions without naming specific values.

\paragraph{GPT-5.1 (7.2K chars).} Identifies wrong \emph{assumptions} rather than wrong behaviors. Names core patterns explicitly (e.g., ``User eligible but no change/cancel requested, core of this task''). Fully parameterized with no scenario-specific values.

\section{Cheatsheet Examples}
\label{app:cheatsheet}

Figures~\ref{fig:cheatsheet-gpt4o} and~\ref{fig:cheatsheet-gpt51} show representative cheatsheets from the weakest and strongest reflection models on \taubench{} scenarios. Figure~\ref{fig:cheatsheet-bfcl} shows a BFCL example. GPT-4o produces short, general guidance (1.8K chars); GPT-5.1 produces detailed, parameterized procedures (7--9K chars) with explicit step numbering, tool call patterns, and state tracking rules.

\begin{figure*}[t]
\small
\begin{verbatim}
## TASK CHEATSHEET
## Handling Exchanges and Cancellations

<failures>
- Attempting to exchange items not marked as delivered.
- Multiple tool calls executed simultaneously.
- Failure to confirm refund and payment options with user before proceeding.
</failures>

<guidance>
   <steps>
   - Authenticate the user first using their email, or name and zip code.
   - Verify the status of each order pertaining to the request.
   - Ensure the items for exchange are part of a delivered order.
   - Collect all necessary details including user confirmation for each task.
   - For exchanges, confirm items to be exchanged and check availability.
   - Ensure single tool call execution at a time.
   - Confirm with the user their choice of payment method before processing.
   - List all intended actions and gain user confirmation before executing.
   </steps>
</guidance>
\end{verbatim}
\caption{Full cheatsheet generated by GPT-4o (self-reflection) for a \taubench{} retail exchange scenario. The guidance is general and mixes procedural advice with policy reminders. Discovery instructions are present (``verify the status,'' ``check availability'') but brief.}
\label{fig:cheatsheet-gpt4o}
\end{figure*}

\begin{figure*}[t]
\small
\begin{verbatim}
## TASK CHEATSHEET
## MMS Issues When User Is Abroad, With Possible Data Limit,
## Roaming, and Device Misconfiguration

<failures>
- Focusing only on device-side MMS troubleshooting and ignoring
  account-level causes (data limit reached, roaming disabled).
- Selecting the wrong line in a multi-line account.
- Transferring to human agent before exhausting all steps.
</failures>

<guidance>
   <steps>
   - 1. Identify the customer and the correct line
     - Call get_customer_by_phone(phone_number=<user_phone>).
     - For each line_id, call get_details_by_id(id=<line_id>).
     - Select the line where phone_number matches the user's number.
   - 2. Gather plan and usage information
     - Call get_details_by_id(id=<plan_id>) for data_limit_gb.
     - Call get_data_usage(customer_id=..., line_id=...).
   - 3. Run basic MMS/device connectivity checks
     - If status bar shows 2G, set_network_mode_preference("4g_5g").
   - 4. Check Wi-Fi Calling (toggle off if ON)
   - 5. Verify messaging app permissions (grant sms, storage)
   - 6. Handle data limit (refuel_data up to 2GB with cost confirm)
   - 7. Handle roaming (enable_roaming + device toggle_roaming)
   - 8. Test can_send_mms() after each major change
   - 9. Tool-call discipline: one call per turn, no mixing
   - 10. Transfer to human only after exhausting all steps
   </steps>
</guidance>
\end{verbatim}
\caption{Abbreviated cheatsheet generated by GPT-5.1 for a \taubench{} telecom MMS scenario (full version: 8.8K chars). The guidance is parameterized (``\texttt{<user\_phone>}'', ``\texttt{<line\_id>}'') with no hardcoded IDs. Each step specifies which tool to call and what to do with its return value.}
\label{fig:cheatsheet-gpt51}
\end{figure*}

\begin{figure*}[t]
\small
\begin{verbatim}
## TASK CHEATSHEET

<failures>
- The model tried to start the engine immediately after converting liters
  to gallons, but VehicleControlAPI required prerequisite state changes.
- The model missed required steps: locking all doors, pressing the brake
  pedal, then starting the engine.
- The model used the healthy_tire_pressure field from check_tire_pressure,
  but the task-defined rule must override: healthy means all four tire
  pressures are between 32 and 35 inclusive.
</failures>

<guidance>
<steps>
1. If the user does not provide the liter amount, ask for it first.
2. Once provided, call liter_to_gallon({"liter": <number>}).
3. For VehicleControlAPI, perform the required setup sequence:
   - lockDoors({"unlock": false,
       "door": ["driver","passenger","rear_left","rear_right"]})
   - pressBrakePedal({"pedalPosition": 1})
   - startEngine({"ignitionMode": "START"})
4. For tire pressure: call check_tire_pressure({}).
   - Healthy only if every tire is between 32 and 35 inclusive.
5. For tweet: post_tweet({"content":"healthy"|"not healthy",
     "tags":["#CarMaintenance"],"mentions":["@VehicleGuru"]})
6. For retweet: use the tweet id returned by post_tweet.
</steps>
</guidance>
\end{verbatim}
\caption{Abbreviated BFCL cheatsheet (GPT-5.4 teacher, \texttt{miss\_param} category). The cheatsheet teaches API prerequisite sequencing (lock doors before engine start) and evaluation-specific rules (tire health threshold). All argument values are schema constants or user-provided; no task-specific IDs are hardcoded.}
\label{fig:cheatsheet-bfcl}
\end{figure*}

\section{Hyperparameters}
\label{app:hyperparams}

Both models use the same QLoRA configuration.

\begin{table}[h]
\centering
\small
\begin{tabular}{ll}
\toprule
\textbf{Parameter} & \textbf{Value} \\
\midrule
LoRA rank & 32 \\
LoRA alpha & 32 \\
Quantization & 4-bit (QLoRA) \\
Learning rate & 2e-4 \\
LR scheduler & cosine \\
Warmup steps & 20 \\
Epochs & 3 \\
Batch size & 4 \\
Max sequence length & 16{,}384 \\
Weight decay & 0.1 \\
\bottomrule
\end{tabular}
\caption{QLoRA training hyperparameters (shared across both models).}
\label{tab:hyperparams}
\end{table}

\section{Recovery Breakdown}
\label{app:recovery-breakdown}

Per-scenario optimization recovery includes two sources: \emph{improved} scenarios where
the cheatsheet enabled the teacher to pass (initial\_score $< 1$,
best\_score $\geq 1$), and \emph{replay} scenarios where the teacher
passed on a stochastic re-execution without a cheatsheet
(initial\_score $\geq 1$). Table~\ref{tab:recovery-breakdown} reports
the breakdown for the 82-task cross-domain reflector study.

\begin{table}[ht]
\centering
\small
\resizebox{\columnwidth}{!}{%
\begin{tabular}{llcccc}
\toprule
\textbf{Reflector} & \textbf{Domain} & \textbf{Improved} & \textbf{Replay} & \textbf{Total} & \textbf{Rate} \\
\midrule
\multirow{4}{*}{GPT-4o}
  & airline  & 4 & 1 & 5/13  & 38.5\% \\
  & retail   & 9 & 2 & 11/14 & 78.6\% \\
  & telecom  & 3 & 1 & 4/55  & 7.3\%  \\
  & overall  & 16 & 4 & 20/82 & 24.4\% \\
\midrule
\multirow{4}{*}{GPT-5-mini}
  & airline  & 9 & 1 & 10/13 & 76.9\% \\
  & retail   & 9 & 3 & 12/14 & 85.7\% \\
  & telecom  & 27 & 0 & 27/55 & 49.1\% \\
  & overall  & 45 & 4 & 49/82 & 59.8\% \\
\midrule
\multirow{4}{*}{GPT-4.1}
  & airline  & 8 & 2 & 10/13 & 76.9\% \\
  & retail   & 9 & 5 & 14/14 & 100\%  \\
  & telecom  & 44 & 0 & 44/55 & 80.0\% \\
  & overall  & 61 & 7 & 68/82 & 82.9\% \\
\midrule
\multirow{4}{*}{GPT-5.1}
  & airline  & 12 & 0 & 12/13 & 92.3\% \\
  & retail   & 13 & 1 & 14/14 & 100\%  \\
  & telecom  & 52 & 2 & 54/55 & 98.2\% \\
  & overall  & 77 & 3 & 80/82 & 97.6\% \\
\bottomrule
\end{tabular}%
}
\caption{Recovery breakdown for the 82-task reflector ablation.
\emph{Improved}: cheatsheet enabled the teacher to pass.
\emph{Replay}: teacher passed on stochastic re-execution without a
cheatsheet. Replay accounts for 4--7 of the total recoveries per
reflector; the majority of recoveries are cheatsheet-driven.}
\label{tab:recovery-breakdown}
\end{table}

\section{Missing Functions (\texttt{miss\_func})}
\label{app:miss-func}

BFCL v4 multi-turn includes a \texttt{miss\_func} category that tests
two capabilities~\citep{pmlr-v267-patil25a}: (1)~the model must identify
that no available function can fulfill a user request, and (2)~once the
missing functions are provided in the next turn, the model must use
them. Functions are held out from the tool list at the start and
silently added back at a specific ``holdout turn,'' where the user says
``I have updated some more functions you can choose from. What about
now?''

This is the hardest category for all models tested. The pass rates here
are measured over the full 200-task \texttt{miss\_func} category (this
deep-dive uses the full category for statistical power; the headline
per-category results in Table~\ref{tab:bfcl-categories} use the 60-task
test split): GPT-4o passes 43\%, Qwen3-4B-Instruct-2507 passes 10.5\%
(21 of 200), and Gemma~4 E4B-it passes 0.5\% (1 of 200). On the 60-task
test split, the corresponding baselines are 11.7\% (Qwen) and 0.0\%
(Gemma; Table~\ref{tab:bfcl-categories}).

\paragraph{Where does Gemma fail?}
We analyzed 50 tasks from the full category to determine which step causes failure. On the
turn \emph{before} the holdout (when the requested function is
missing), Gemma uses a different tool instead of indicating inability
in 72\% of cases (36/50). On the holdout turn itself (when the missing
function is added back), Gemma fails to call the newly available
function in 96\% of cases (48/50), producing a text response instead
of a tool call.

By contrast, GPT-4o also uses a different tool on the pre-holdout turn
(96\%, 47/49), but succeeds on the holdout turn 76\% of the time
(37/49). Both models rarely indicate inability on the pre-holdout turn;
the primary differentiator is whether the model calls the newly added
function when prompted.

\paragraph{Offline replay.}
We replayed the holdout turn for \texttt{miss\_func\_1} using the
\texttt{google/gemma-4-E4B-it} checkpoint locally with the same chat
template used in our vLLM evaluation pipeline. We verified from the
BFCL inference logs that the tool list grows from 17 to 18 functions on
the holdout turn, with \texttt{mv} appended. The model's system prompt
includes all 18 tools in \texttt{<|tool>} declarations. The user
message on the holdout turn is: ``I have updated some more functions you
can choose from. What about now?'' It does not name which function was
added. The model produces:

\begin{quote}
\small
``I'm ready for the new functions! Please provide them, and I'll be
happy to see what I can do with them.''
\end{quote}

The model generates text instead of a tool call.

\paragraph{Ablation: explicit vs.\ vague prompt.}
To test whether the failure is due to the model's inability to use newly
added tools or the vagueness of the holdout prompt, we replaced the
BFCL default message (``I have updated some more functions you can
choose from. What about now?'') with an explicit instruction: ``I have
added a sort function to your available tools. Please sort the file
now.'' With this change, the model correctly produces a tool call to
the newly added function. This confirms that Gemma can use tools added
mid-conversation when the instruction names the function and the
action; the failure on \texttt{miss\_func} is specific to the vague
holdout prompt, which does not indicate which function was added or
what action to take.

\paragraph{Effect of fine-tuning.}
After fine-tuning with combined data, Gemma reaches 21.5\% on
\texttt{miss\_func} (43 of 200). The fine-tuned model learned to
respond with tool calls on holdout turns rather than text, reducing the
text-only response rate from 85.5\% to 22.0\% of turns on this
category.

\section{Trajectory Communication Patterns}
\label{app:trajectory-patterns}

We present text responses from GPT-4o and GPT-5.1 (high reasoning) on the same task, both without any cheatsheet scaffold (replay passes). Both models have just called \texttt{get\_details\_by\_id}, received structured account information confirming the line is active and the issue is phone-side, and now communicate next steps to the user. The task is MMS troubleshooting where the user's mobile data is off, Wi-Fi calling interferes, and the SIM is unseated.

\paragraph{GPT-4o, no cheatsheet (305 characters).}

\begin{quote}
\small
I've found your account and confirmed that your line is active. Let's proceed with troubleshooting the MMS issue. Could you please check if your phone has cellular service and mobile data enabled? You can do this by checking the status bar on your phone for signal strength and data connection indicators.
\end{quote}

\paragraph{GPT-5.1 high reasoning, no cheatsheet (1,132 characters).}

\begin{quote}
\small
I've pulled up your line ending in 2002. Your line is active, roaming is enabled, and you still have data left in your plan, so nothing on the account/plan side is blocking MMS. Let's check the phone side next.

First I want to confirm that \textbf{mobile data itself is working}, since MMS depends on that.

Please do this and tell me the results:

\textbf{1. Turn off Wi-Fi completely}\\
Swipe down from the top and tap the Wi-Fi icon so it's OFF.

\textbf{2. Make sure Mobile Data is ON}\\
In the same quick settings area, make sure ``Mobile data'' is turned ON.

\textbf{3. Test mobile data}\\
With Wi-Fi still OFF, open your browser and try to load any website.

\textbf{4. Tell me:}\\
Does the website load or not? When you try to send a picture message right now (with Wi-Fi still OFF and mobile data ON), does it stay stuck on ``sending,'' show a specific error message, or fail immediately?

Once you tell me what happens in steps 3 and 4, I'll guide you through the next specific checks (network type/APN/Wi-Fi Calling/app permissions) based on your result.
\end{quote}

\paragraph{Observations.} Both models have received the same account information from \texttt{get\_details\_by\_id} and both intend the same next action: verify that the user's mobile data connection works. GPT-4o makes a single request and waits (305 characters). GPT-5.1 first summarizes what it learned from the tool return, then provides a numbered protocol with specific UI instructions and a conditional branching plan for what comes next (1,132 characters; 3.7$\times$ longer).

This illustrates the interaction style difference reported in Section~\ref{subsec:response-complexity}: GPT-4o sequences diagnostics one at a time across turns (incremental), while GPT-5.1 front-loads a multi-step plan into a single turn (batch). Both strategies resolve the task successfully (reward = 1.0). The difference is not in the number of tool calls (comparable on average across the matched set) but in how much context each model provides per text turn.

\paragraph{Task and provenance.} Task ID: {\small\texttt{[mms\_issue]\allowbreak bad\_wifi\_calling|\allowbreak data\_mode\_off|\allowbreak unseat\_sim\_card\allowbreak [PERSONA:Easy]}}. Both trajectories are from independent generation runs on the same task definition without cheatsheet guidance (improvement\_source = replay\_only). This example is near the median of the length ratio distribution (3.7$\times$ vs.\ 3.8$\times$ median on 259 matched tasks).

\paragraph{GPT-4o with cheatsheet vs.\ GPT-5.1 direct.}

The example above uses replay-only passes (no cheatsheet) to isolate intrinsic model style. We now show that GPT-4o with cheatsheet guidance exhibits the same incremental pattern. The following responses are from a cheatsheet-recovered task in the 259-task matched set, both after calling \texttt{get\_customer\_by\_phone} and identifying the user's line.

\subparagraph{GPT-4o + cheatsheet (299 characters).}

\begin{quote}
\small
Thank you for the information. I've identified your line associated with the phone number 555-123-2002. Let's proceed with troubleshooting the MMS issue.

Could you please check if your phone can send MMS messages by running the \texttt{can\_send\_mms()} diagnostic on your device and let me know the result?
\end{quote}

\subparagraph{GPT-5.1 direct, no cheatsheet (1,141 characters).}

\begin{quote}
\small
Please try them now, one by one. That will give us clear answers:

\textbf{1. Turn off Wi-Fi first} so we're only testing mobile data.\\
\textbf{2. Then, in this order, tell me what happens for each:}
\begin{itemize}[nosep,leftmargin=1em]
\item Regular call: Try calling any number. Does it connect and stay connected?
\item Regular SMS text (no picture): Send a plain text to someone. Does it send?
\item Internet over mobile data: With Wi-Fi still OFF, open a browser. Does it load normally, slowly, or not at all?
\item MMS (picture message): Try sending a photo. Does it stay on ``sending,'' show ``failed,'' or something else?
\end{itemize}

Also, please look at the top of your screen and tell me what network type/icon you see (e.g., 4G, LTE, 5G, E, H, R) and whether there's any roaming icon.

Once you report back with these results, I'll run checks on your line and walk you through the next steps.
\end{quote}

\subparagraph{Observations.} GPT-4o with cheatsheet produces the same incremental pattern as GPT-4o without cheatsheet: identify the line, request one diagnostic, wait (299 characters). GPT-5.1 requests a full test battery covering calls, SMS, data, and MMS in a single turn (1,141 characters; 3.8$\times$ longer). The cheatsheet specifies diagnostic ordering (which tools to call and in what sequence), not response verbosity or framing; GPT-4o's communication style remains unchanged even on tasks it originally failed. Task ID: {\small\texttt{[mms\_issue]\allowbreak airplane\_mode\_on|...|data\_usage\_exceeded|\allowbreak user\_abroad\_roaming\_enabled\_off\allowbreak [PERSONA:None]}}.

\paragraph{Per-tool response length breakdown.}

The qualitative examples above use replay-only and cheatsheet-guided trajectories to illustrate the style difference. Table~\ref{tab:per-tool-length} below reports aggregate statistics on the full 259-task matched set. The ratios are consistent across tool types and across both conditions, confirming that the cheatsheet does not alter GPT-4o's communication pattern.

\begin{table}[ht]
\centering
\small
\resizebox{\columnwidth}{!}{%
\begin{tabular}{lccc}
\toprule
\textbf{After tool} & \textbf{GPT-4o} & \textbf{GPT-5.1} & \textbf{Ratio} \\
 & \textbf{(+cheatsheet)} & \textbf{(direct)} & \\
\midrule
\texttt{get\_details\_by\_id} & 469 & 997 & 2.1$\times$ \\
\texttt{enable\_roaming} & 292 & 1,254 & 4.3$\times$ \\
\texttt{get\_data\_usage} & 476 & 1,450 & 3.0$\times$ \\
\texttt{refuel\_data} & 206 & 835 & 4.1$\times$ \\
\midrule
All text responses & 308 & 987 & 3.2$\times$ \\
\bottomrule
\end{tabular}%
}
\caption{Average text response length (characters) after each tool call type on 259 matched tasks. The ratio is consistent across tool types (2.1--4.3$\times$), confirming the style difference is systematic rather than specific to particular actions.}
\label{tab:per-tool-length}
\end{table}

\section{Student Model Comparison}
\label{app:model-comparison}

Table~\ref{tab:model-comparison} summarizes the two student models. Both operate at comparable effective scale (${\sim}$4B parameters) but differ in architecture, modality, and tool-call format. Training on both and observing consistent improvements across the two confirms that the method generalizes across model families.

\begin{table}[t]
\centering
\small
\begin{tabular}{ll}
\toprule
\multicolumn{2}{c}{\textbf{Gemma 4 E4B-it}} \\
\midrule
Parameters   & 4.5B effective (8B total) \\
Architecture & Multimodal (text, image, audio) \\
Layers / Hidden & 42 / 2{,}560 \\
Attention    & 8Q / 2KV, hybrid sliding+full \\
Context      & 128K tokens \\
Tool format  & Gemma special tokens \\
\midrule
\multicolumn{2}{c}{\textbf{Qwen3-4B-Instruct-2507}} \\
\midrule
Parameters   & 4.0B (3.6B non-emb.) \\
Architecture & Text-only causal LM \\
Layers / Hidden & 36 / 2{,}560 \\
Attention    & 32Q / 8KV, full \\
Context      & 262K tokens \\
Tool format  & JSON in XML tags \\
\bottomrule
\end{tabular}
\caption{Student model comparison. Both share a hidden dimension of
2{,}560 and operate at comparable effective scale but differ in
architecture, modality, and tool-call format.}
\label{tab:model-comparison}
\end{table}

\section{Training Data Composition}
\label{app:data-composition}

Table~\ref{tab:data-composition} reports the training data composition for both benchmarks. Each multi-turn trajectory is unrolled into prompt/completion pairs (one per assistant turn).

\begin{table}[t]
\centering
\small
\resizebox{\columnwidth}{!}{%
\begin{tabular}{lccc}
\toprule
 & \textbf{Filtered} & \textbf{Recovered} & \textbf{Combined} \\
\midrule
\multicolumn{4}{l}{\textit{\taubench{} telecom (2{,}171 non-base tasks)}} \\
\quad Trajectories & 158 & 279 & 437 \\
\quad Unrolled examples & 2{,}049 & 4{,}814 & 6{,}863 \\
\quad Recovered:filtered & \multicolumn{3}{c}{1.77:1} \\
\midrule
\multicolumn{4}{l}{\textit{BFCL v4 multi-turn (560 train tasks)}} \\
\quad base & 96 & 11 & 107 \\
\quad long\_context & 78 & 20 & 98 \\
\quad miss\_func & 66 & 22 & 88 \\
\quad miss\_param & 56 & 36 & 92 \\
\cmidrule{2-4}
\quad Total trajectories & 296 & 89 & 385 \\
\quad Unrolled examples & 1{,}773 & 954 & 2{,}727 \\
\quad Recovered:filtered & \multicolumn{3}{c}{0.30:1} \\
\bottomrule
\end{tabular}%
}
\caption{Training data composition. Filtered = tasks the teacher passed.
Recovered = tasks recovered via per-scenario optimization. For BFCL, the
264 failures (560 train tasks minus 296 passed) yield 89 recoveries, a
33.7\% recovery rate.}
\label{tab:data-composition}
\end{table}

\section{BFCL Turn-Level Partial Reward}
\label{app:bfcl-partial-reward}

BFCL v4 multi-turn evaluates tasks as binary pass/fail (all turns must
pass for the task to succeed). The benchmark does not provide a built-in
partial reward metric. We implement turn-level partial reward as the
average fraction of turns passed per task:
\[
\text{Partial Reward}(t) = \frac{\text{turns passed in task } t}{\text{total turns in task } t}
\]
This serves as a per-turn accuracy proxy, analogous to the action
accuracy component of \taubench{}'s reward decomposition, allowing us to
measure whether fine-tuning improves tool-calling behavior at the turn
level even when end-to-end task accuracy is unchanged.

\section{BFCL Per-Category Results}
\label{app:bfcl-categories}

Table~\ref{tab:bfcl-categories} reports per-category accuracy on BFCL v4
multi-turn, averaged across 3 evaluation runs.

\begin{table}[ht]
\centering
\small
\resizebox{\columnwidth}{!}{%
\begin{tabular}{lcccc}
\toprule
 & \multicolumn{2}{c}{\textbf{Gemma 4 E4B-it}} & \multicolumn{2}{c}{\textbf{Qwen3-4B-Inst.}} \\
\cmidrule(lr){2-3} \cmidrule(lr){4-5}
\textbf{Category} & Base & +Comb. & Base & +Comb. \\
\midrule
base           & 17.2 & 23.3 & 30.0 & 28.9 \\
long\_context  & 15.0 & 18.3 & 28.3 & 21.1 \\
miss\_func     & 0.0  & \textbf{15.6} & 11.7 & \textbf{31.7} \\
miss\_param    & 17.8 & 21.7 & 24.4 & 20.6 \\
\midrule
Overall        & 12.5 & \textbf{19.7} & 23.6 & 25.5 \\
\bottomrule
\end{tabular}%
}
\caption{BFCL per-category accuracy (\%, 60 test tasks each, averaged
across 3 runs). Bold indicates improvement over baseline.}
\label{tab:bfcl-categories}
\end{table}

The \texttt{miss\_func} category (tasks requiring the model to recognize
that a requested function is unavailable) is where both models start
lowest (Gemma 0\%, Qwen 11.7\%) and show the clearest improvement
(Gemma 15.6\%, Qwen 31.7\%). This gain is stable across all three
evaluation runs (Appendix~\ref{app:repeated-eval}).

BFCL's turn-level evaluation is strict: each turn has ground-truth
function calls, and if the model produces no decodable tool calls for a
turn, that turn fails immediately regardless of any text response. This
makes partial progress harder to observe than in \taubench{}, where the
verifier evaluates end-state correctness rather than per-turn output
format. Other per-category results show mixed effects at $n{=}60$ and
are exploratory.

\section{BFCL Repeated Evaluation}
\label{app:repeated-eval}

To assess serving non-determinism, we ran three independent evaluations per condition on BFCL v4 multi-turn. Despite deterministic user turns and temp=0, vLLM serving introduces small floating-point differences that flip 1--3 tasks per run.

\begin{table}[t]
\centering
\small
\resizebox{\columnwidth}{!}{%
\begin{tabular}{lcccc}
\toprule
\textbf{Condition} & \textbf{Run 1} & \textbf{Run 2} & \textbf{Run 3} & \textbf{Spread} \\
\midrule
\multicolumn{5}{l}{\textit{Overall accuracy (\%, 240 test tasks)}} \\
\quad Gemma baseline & 12.5 & 12.5 & 12.5 & 0.0pp \\
\quad Gemma + combined & 20.4 & 19.6 & 19.2 & 1.2pp \\
\quad Qwen baseline & 23.8 & 23.8 & 23.3 & 0.5pp \\
\quad Qwen + combined & 25.8 & 26.2 & 24.6 & 1.6pp \\
\midrule
\multicolumn{5}{l}{\textit{Per-category max spread (60 test tasks each)}} \\
\quad Gemma baseline & \multicolumn{4}{c}{1.7pp (base, miss\_param)} \\
\quad Gemma + combined & \multicolumn{4}{c}{3.3pp (long\_context)} \\
\quad Qwen baseline & \multicolumn{4}{c}{1.7pp (miss\_param)} \\
\quad Qwen + combined & \multicolumn{4}{c}{3.3pp (base)} \\
\bottomrule
\end{tabular}%
}
\caption{BFCL v4 repeated evaluation stability. Overall accuracy is stable across runs (spread $\leq$1.6pp). Per-category results at $n{=}60$ vary by up to 3.3pp from serving non-determinism alone. The \texttt{miss\_func} gain is the most stable signal: Gemma 0.0\% $\times$3 (baseline) vs.\ 15.0--16.7\% (combined); Qwen 11.7\% $\times$3 (baseline) vs.\ 31.7\% $\times$3 (combined).}
\label{tab:repeated-eval}
\end{table}

\section{Training Stability Verification}
\label{app:training-stability}

To verify that reported results reflect stable training outcomes rather than favorable random seeds, we conducted independent retraining and re-evaluation runs for our strongest result (Qwen3-4B-Instruct-2507 combined on \taubench{} telecom).

\begin{table}[t]
\centering
\small
\begin{tabular}{llc}
\toprule
\textbf{Run} & \textbf{Description} & \textbf{Pass\textasciicircum{}1} \\
\midrule
Original & Train + eval & 0.529 \\
Re-eval & Same checkpoint, new eval & 0.518 \\
Retrain & New training run + eval & 0.538 \\
\bottomrule
\end{tabular}
\caption{Qwen3-4B-Instruct-2507 combined training stability on \taubench{} telecom (114 tasks, 3 trials, temp=0). Pass\textasciicircum{}1 ranges from 0.518 to 0.538 across independent training and evaluation runs (spread: 0.020), confirming that the reported improvement is not an artifact of a favorable seed.}
\label{tab:training-stability}
\end{table}

The re-evaluation uses the same merged checkpoint served under identical vLLM settings; the difference from the original (0.529 vs.\ 0.518) reflects stochastic user simulation variance. The retrain uses a fresh QLoRA training run with identical hyperparameters; its result (0.538) confirms that the large gain over the baseline (0.132) and filtered-only training (0.111) is stable across independent training and evaluation on the same dataset.

%% file: references.bib
@misc{yuan2023scalingrelationshiplearningmathematical,
      title={Scaling Relationship on Learning Mathematical Reasoning with Large Language Models}, 
      author={Zheng Yuan and Hongyi Yuan and Chengpeng Li and Guanting Dong and Keming Lu and Chuanqi Tan and Chang Zhou and Jingren Zhou},
      year={2023},
      eprint={2308.01825},
      archivePrefix={arXiv},
      primaryClass={cs.CL},
      url={https://arxiv.org/abs/2308.01825}, 
}

@misc{chen2023fireactlanguageagentfinetuning,
      title={{FireAct}: Toward Language Agent Fine-tuning}, 
      author={Baian Chen and Chang Shu and Ehsan Shareghi and Nigel Collier and Karthik Narasimhan and Shunyu Yao},
      year={2023},
      eprint={2310.05915},
      archivePrefix={arXiv},
      primaryClass={cs.CL},
      url={https://arxiv.org/abs/2310.05915}, 
}

@inproceedings{NEURIPS2024_61cce86d,
 author = {Liu, Zuxin and Hoang, Thai and Zhang, Jianguo and Zhu, Ming and Lan, Tian and Kokane, Shirley and Tan, Juntao and Yao, Weiran and Liu, Zhiwei and Feng, Yihao and Murthy, Rithesh and Yang, Liangwei and Savarese, Silvio and Niebles, Juan Carlos and Wang, Huan and Heinecke, Shelby and Xiong, Caiming},
 booktitle = {Advances in Neural Information Processing Systems},
 doi = {10.52202/079017-1725},
 editor = {A. Globerson and L. Mackey and D. Belgrave and A. Fan and U. Paquet and J. Tomczak and C. Zhang},
 pages = {54463--54482},
 publisher = {Curran Associates, Inc.},
 title = {{APIGen}: Automated PIpeline for Generating Verifiable and Diverse Function-Calling Datasets},
 url = {https://proceedings.neurips.cc/paper_files/paper/2024/file/61cce86d180b1184949e58939c4f983d-Paper-Datasets_and_Benchmarks_Track.pdf},
 volume = {37},
 year = {2024}
}

@misc{barres2025tau2benchevaluatingconversationalagents,
      title={$\tau^2$-{Bench}: Evaluating Conversational Agents in a Dual-Control Environment}, 
      author={Victor Barres and Honghua Dong and Soham Ray and Xujie Si and Karthik Narasimhan},
      year={2025},
      eprint={2506.07982},
      archivePrefix={arXiv},
      primaryClass={cs.AI},
      url={https://arxiv.org/abs/2506.07982}, 
}

@InProceedings{pmlr-v267-patil25a,
  title = 	 {The Berkeley Function Calling Leaderboard ({BFCL}): From Tool Use to Agentic Evaluation of Large Language Models},
  author =       {Patil, Shishir G and Mao, Huanzhi and Yan, Fanjia and Ji, Charlie Cheng-Jie and Suresh, Vishnu and Stoica, Ion and Gonzalez, Joseph E.},
  booktitle = 	 {Proceedings of the 42nd International Conference on Machine Learning},
  pages = 	 {48371--48392},
  year = 	 {2025},
  editor = 	 {Singh, Aarti and Fazel, Maryam and Hsu, Daniel and Lacoste-Julien, Simon and Berkenkamp, Felix and Maharaj, Tegan and Wagstaff, Kiri and Zhu, Jerry},
  volume = 	 {267},
  series = 	 {Proceedings of Machine Learning Research},
  month = 	 {13--19 Jul},
  publisher =    {PMLR},
  url = 	 {https://proceedings.mlr.press/v267/patil25a.html}
}

@misc{googledeepmind2026gemma4e4bit,
  title        = {{google/gemma-4-E4B-it}},
  author       = {{Google DeepMind}},
  year         = {2026},
  howpublished = {Hugging Face model card},
  url          = {https://huggingface.co/google/gemma-4-E4B-it},
  note         = {Accessed: 2026-04-28}
}

@misc{yang2025qwen3technicalreport,
      title={{Qwen3} Technical Report}, 
      author={An Yang and Anfeng Li and Baosong Yang and Beichen Zhang and Binyuan Hui and Bo Zheng and Bowen Yu and Chang Gao and Chengen Huang and Chenxu Lv and Chujie Zheng and Dayiheng Liu and Fan Zhou and Fei Huang and Feng Hu and Hao Ge and Haoran Wei and Huan Lin and Jialong Tang and Jian Yang and Jianhong Tu and Jianwei Zhang and Jianxin Yang and Jiaxi Yang and Jing Zhou and Jingren Zhou and Junyang Lin and Kai Dang and Keqin Bao and Kexin Yang and Le Yu and Lianghao Deng and Mei Li and Mingfeng Xue and Mingze Li and Pei Zhang and Peng Wang and Qin Zhu and Rui Men and Ruize Gao and Shixuan Liu and Shuang Luo and Tianhao Li and Tianyi Tang and Wenbiao Yin and Xingzhang Ren and Xinyu Wang and Xinyu Zhang and Xuancheng Ren and Yang Fan and Yang Su and Yichang Zhang and Yinger Zhang and Yu Wan and Yuqiong Liu and Zekun Wang and Zeyu Cui and Zhenru Zhang and Zhipeng Zhou and Zihan Qiu},
      year={2025},
      eprint={2505.09388},
      archivePrefix={arXiv},
      primaryClass={cs.CL},
      url={https://arxiv.org/abs/2505.09388}, 
}

@inproceedings{3666122.3666563,
author = {Dettmers, Tim and Pagnoni, Artidoro and Holtzman, Ari and Zettlemoyer, Luke},
title = {{QLoRA}: efficient finetuning of quantized {LLM}s},
year = {2023},
publisher = {Curran Associates Inc.},
address = {Red Hook, NY, USA},
booktitle = {Proceedings of the 37th International Conference on Neural Information Processing Systems},
articleno = {441},
numpages = {28},
location = {New Orleans, LA, USA},
series = {NIPS '23}
}

@misc{agrawal2026gepareflectivepromptevolution,
      title={{GEPA}: Reflective Prompt Evolution Can Outperform Reinforcement Learning}, 
      author={Lakshya A Agrawal and Shangyin Tan and Dilara Soylu and Noah Ziems and Rishi Khare and Krista Opsahl-Ong and Arnav Singhvi and Herumb Shandilya and Michael J Ryan and Meng Jiang and Christopher Potts and Koushik Sen and Alexandros G. Dimakis and Ion Stoica and Dan Klein and Matei Zaharia and Omar Khattab},
      year={2026},
      eprint={2507.19457},
      archivePrefix={arXiv},
      primaryClass={cs.CL},
      url={https://arxiv.org/abs/2507.19457}, 
}

@inproceedings{pryzant-etal-2023-automatic,
    title = "Automatic Prompt Optimization with ``Gradient Descent'' and Beam Search",
    author = "Pryzant, Reid  and
      Iter, Dan  and
      Li, Jerry  and
      Lee, Yin  and
      Zhu, Chenguang  and
      Zeng, Michael",
    editor = "Bouamor, Houda  and
      Pino, Juan  and
      Bali, Kalika",
    booktitle = "Proceedings of the 2023 Conference on Empirical Methods in Natural Language Processing",
    month = dec,
    year = "2023",
    address = "Singapore",
    publisher = "Association for Computational Linguistics",
    url = "https://aclanthology.org/2023.emnlp-main.494/",
    doi = "10.18653/v1/2023.emnlp-main.494",
    pages = "7957--7968"
}

@inproceedings{ICLR2024_3339f19c,
 author = {Yang, Chengrun and Wang, Xuezhi and Lu, Yifeng and Liu, Hanxiao and Le, Quoc V and Zhou, Denny and Chen, Xinyun},
 booktitle = {International Conference on Learning Representations},
 editor = {B. Kim and Y. Yue and S. Chaudhuri and K. Fragkiadaki and M. Khan and Y. Sun},
 pages = {12028--12068},
 title = {Large Language Models as Optimizers},
 url = {https://proceedings.iclr.cc/paper_files/paper/2024/file/3339f19c5fcee3ad74502947a32be9e6-Paper-Conference.pdf},
 volume = {2024},
 year = {2024}
}

@inproceedings{opsahl-ong-etal-2024-optimizing,
    title = "Optimizing Instructions and Demonstrations for Multi-Stage Language Model Programs",
    author = "Opsahl-Ong, Krista  and
      Ryan, Michael J  and
      Purtell, Josh  and
      Broman, David  and
      Potts, Christopher  and
      Zaharia, Matei  and
      Khattab, Omar",
    editor = "Al-Onaizan, Yaser  and
      Bansal, Mohit  and
      Chen, Yun-Nung",
    booktitle = "Proceedings of the 2024 Conference on Empirical Methods in Natural Language Processing",
    month = nov,
    year = "2024",
    address = "Miami, Florida, USA",
    publisher = "Association for Computational Linguistics",
    url = "https://aclanthology.org/2024.emnlp-main.525/",
    doi = "10.18653/v1/2024.emnlp-main.525",
    pages = "9340--9366"
}

@misc{yuan2025agentrtraininglanguagemodel,
      title={{Agent-R}: Training Language Model Agents to Reflect via Iterative Self-Training}, 
      author={Siyu Yuan and Zehui Chen and Zhiheng Xi and Junjie Ye and Zhengyin Du and Jiecao Chen},
      year={2025},
      eprint={2501.11425},
      archivePrefix={arXiv},
      primaryClass={cs.AI},
      url={https://arxiv.org/abs/2501.11425}, 
}

@inproceedings{ICLR2024_28e50ee5,
 author = {Qin, Yujia and Liang, Shihao and Ye, Yining and Zhu, Kunlun and Yan, Lan and Lu, Yaxi and Lin, Yankai and Cong, Xin and Tang, Xiangru and Qian, Bill and Zhao, Sihan and Hong, Lauren and Tian, Runchu and Xie, Ruobing and Zhou, Jie and Gerstein, Mark and li, dahai and Liu, Zhiyuan and Sun, Maosong},
 booktitle = {International Conference on Learning Representations},
 editor = {B. Kim and Y. Yue and S. Chaudhuri and K. Fragkiadaki and M. Khan and Y. Sun},
 pages = {9695--9717},
 title = {{ToolLLM}: Facilitating Large Language Models to Master 16000+ Real-world {API}s},
 url = {https://proceedings.iclr.cc/paper_files/paper/2024/file/28e50ee5b72e90b50e7196fde8ea260e-Paper-Conference.pdf},
 volume = {2024},
 year = {2024}
}

@inproceedings{zhang-etal-2025-xlam,
    title = "x{LAM}: A Family of Large Action Models to Empower {AI} Agent Systems",
    author = "Zhang, Jianguo  and
      Lan, Tian  and
      Zhu, Ming  and
      Liu, Zuxin  and
      Hoang, Thai  and
      Kokane, Shirley  and
      Yao, Weiran  and
      Tan, Juntao  and
      Liu, Zhiwei  and
      Feng, Yihao  and
      Niebles, Juan Carlos  and
      Heinecke, Shelby  and
      Wang, Huan  and
      Savarese, Silvio  and
      Xiong, Caiming",
    editor = "Chiruzzo, Luis  and
      Ritter, Alan  and
      Wang, Lu",
    booktitle = "Proceedings of the 2025 Conference of the Nations of the Americas Chapter of the Association for Computational Linguistics: Human Language Technologies (Volume 1: Long Papers)",
    month = apr,
    year = "2025",
    address = "Albuquerque, New Mexico",
    publisher = "Association for Computational Linguistics",
    url = "https://aclanthology.org/2025.naacl-long.578/",
    doi = "10.18653/v1/2025.naacl-long.578",
    pages = "11583--11597",
    ISBN = "979-8-89176-189-6"
}

@inproceedings{yin-etal-2025-magnet,
    title = "Magnet: Multi-turn Tool-use Data Synthesis and Distillation via Graph Translation",
    author = "Yin, Fan  and
      Wang, Zifeng  and
      Hsu, I-Hung  and
      Yan, Jun  and
      Jiang, Ke  and
      Chen, Yanfei  and
      Gu, Jindong  and
      Le, Long  and
      Chang, Kai-Wei  and
      Lee, Chen-Yu  and
      Palangi, Hamid  and
      Pfister, Tomas",
    booktitle = "Proceedings of the 63rd Annual Meeting of the Association for Computational Linguistics (Volume 1: Long Papers)",
    month = jul,
    year = "2025",
    address = "Vienna, Austria",
    publisher = "Association for Computational Linguistics",
    url = "https://aclanthology.org/2025.acl-long.1566/",
    doi = "10.18653/v1/2025.acl-long.1566",
    pages = "32600--32616",
}

@article{
singh2024beyond,
title={Beyond Human Data: Scaling Self-Training for Problem-Solving with Language Models},
author={Avi Singh and John D Co-Reyes and Rishabh Agarwal and Ankesh Anand and Piyush Patil and Xavier Garcia and Peter J Liu and James Harrison and Jaehoon Lee and Kelvin Xu and Aaron T Parisi and Abhishek Kumar and Alexander A Alemi and Alex Rizkowsky and Azade Nova and Ben Adlam and Bernd Bohnet and Gamaleldin Fathy Elsayed and Hanie Sedghi and Igor Mordatch and Isabelle Simpson and Izzeddin Gur and Jasper Snoek and Jeffrey Pennington and Jiri Hron and Kathleen Kenealy and Kevin Swersky and Kshiteej Mahajan and Laura A Culp and Lechao Xiao and Maxwell Bileschi and Noah Constant and Roman Novak and Rosanne Liu and Tris Warkentin and Yamini Bansal and Ethan Dyer and Behnam Neyshabur and Jascha Sohl-Dickstein and Noah Fiedel},
journal={Transactions on Machine Learning Research},
issn={2835-8856},
year={2024},
url={https://openreview.net/forum?id=lNAyUngGFK},
note={Expert Certification}
}

@inproceedings{hosseini2024vstar,
title     = {{V-STaR}: Training Verifiers for Self-Taught Reasoners},
author    = {Hosseini, Arian and Yuan, Xingdi and Malkin, Nikolay and Courville, Aaron and Sordoni, Alessandro and Agarwal, Rishabh},
booktitle = {Proceedings of the 2024 Conference on Language Modeling},
year      = {2024},
month     = jul,
url       = {https://colmweb.org/2024/AcceptedPapers.html}
}

@inproceedings{song-etal-2024-trial,
    title = "Trial and Error: Exploration-Based Trajectory Optimization of {LLM} Agents",
    author = "Song, Yifan  and
      Yin, Da  and
      Yue, Xiang  and
      Huang, Jie  and
      Li, Sujian  and
      Lin, Bill Yuchen",
    editor = "Ku, Lun-Wei  and
      Martins, Andre  and
      Srikumar, Vivek",
    booktitle = "Proceedings of the 62nd Annual Meeting of the Association for Computational Linguistics (Volume 1: Long Papers)",
    month = aug,
    year = "2024",
    address = "Bangkok, Thailand",
    publisher = "Association for Computational Linguistics",
    url = "https://aclanthology.org/2024.acl-long.409/",
    doi = "10.18653/v1/2024.acl-long.409",
    pages = "7584--7600"
}

@inproceedings{3666122.3666499,
author = {Shinn, Noah and Cassano, Federico and Gopinath, Ashwin and Narasimhan, Karthik and Yao, Shunyu},
title = {Reflexion: language agents with verbal reinforcement learning},
year = {2023},
publisher = {Curran Associates Inc.},
address = {Red Hook, NY, USA},
booktitle = {Proceedings of the 37th International Conference on Neural Information Processing Systems},
articleno = {377},
numpages = {19},
location = {New Orleans, LA, USA},
series = {NIPS '23}
}

@inproceedings{ICLR2024_f1cf02ce,
 author = {Khattab, Omar and Singhvi, Arnav and Maheshwari, Paridhi and Zhang, Zhiyuan and Santhanam, Keshav and A, Sri Vardhamanan and Haq, Saiful and Sharma, Ashutosh and Joshi, Thomas and Moazam, Hanna and Miller, Heather and Zaharia, Matei and Potts, Christopher},
 booktitle = {International Conference on Learning Representations},
 editor = {B. Kim and Y. Yue and S. Chaudhuri and K. Fragkiadaki and M. Khan and Y. Sun},
 pages = {54928--54958},
 title = {{DSPy}: Compiling Declarative Language Model Calls into State-of-the-Art Pipelines},
 url = {https://proceedings.iclr.cc/paper_files/paper/2024/file/f1cf02ce09757f57c3b93c0db83181e0-Paper-Conference.pdf},
 volume = {2024},
 year = {2024}
}

@inproceedings{ICLR2024_82eec786,
 author = {Xu, Can and Sun, Qingfeng and Zheng, Kai and Geng, Xiubo and Zhao, Pu and Feng, Jiazhan and Tao, Chongyang and Lin, Qingwei and Jiang, Daxin},
 booktitle = {International Conference on Learning Representations},
 editor = {B. Kim and Y. Yue and S. Chaudhuri and K. Fragkiadaki and M. Khan and Y. Sun},
 pages = {30745--30766},
 title = {{WizardLM}: Empowering Large Pre-Trained Language Models to Follow Complex Instructions},
 url = {https://proceedings.iclr.cc/paper_files/paper/2024/file/82eec786fdfbbfa53450c5feb7d1ac92-Paper-Conference.pdf},
 volume = {2024},
 year = {2024}
}

@misc{zhang2026agenticcontextengineeringevolving,
      title={Agentic Context Engineering: Evolving Contexts for Self-Improving Language Models}, 
      author={Qizheng Zhang and Changran Hu and Shubhangi Upasani and Boyuan Ma and Fenglu Hong and Vamsidhar Kamanuru and Jay Rainton and Chen Wu and Mengmeng Ji and Hanchen Li and Urmish Thakker and James Zou and Kunle Olukotun},
      year={2026},
      eprint={2510.04618},
      archivePrefix={arXiv},
      primaryClass={cs.LG},
      url={https://arxiv.org/abs/2510.04618}, 
}

@misc{zhai2026doesrlexpandcapability,
      title={Does {RL} Expand the Capability Boundary of {LLM} Agents? A {PASS@(k,T)} Analysis}, 
      author={Zhiyuan Zhai and Wenjing Yan and Xiaodan Shao and Xin Wang},
      year={2026},
      eprint={2604.14877},
      archivePrefix={arXiv},
      primaryClass={cs.LG},
      url={https://arxiv.org/abs/2604.14877}, 
}

@inproceedings{3600270.3601396,
author = {Zelikman, Eric and Wu, Yuhuai and Mu, Jesse and Goodman, Noah D.},
title = {{STaR}: self-taught reasoner bootstrapping reasoning with reasoning},
year = {2022},
isbn = {9781713871088},
publisher = {Curran Associates Inc.},
address = {Red Hook, NY, USA},
booktitle = {Proceedings of the 36th International Conference on Neural Information Processing Systems},
articleno = {1126},
numpages = {13},
location = {New Orleans, LA, USA},
series = {NIPS '22}
}

@misc{gulcehre2023reinforcedselftrainingrestlanguage,
      title={Reinforced Self-Training ({ReST}) for Language Modeling}, 
      author={Caglar Gulcehre and Tom Le Paine and Srivatsan Srinivasan and Ksenia Konyushkova and Lotte Weerts and Abhishek Sharma and Aditya Siddhant and Alex Ahern and Miaosen Wang and Chenjie Gu and Wolfgang Macherey and Arnaud Doucet and Orhan Firat and Nando de Freitas},
      year={2023},
      eprint={2308.08998},
      archivePrefix={arXiv},
      primaryClass={cs.CL},
      url={https://arxiv.org/abs/2308.08998}, 
}

@inproceedings{ICLR2024_5be69a58,
 author = {Agarwal, Rishabh and Vieillard, Nino and Zhou, Yongchao and Stanczyk, Piotr and Ramos Garea, Sabela and Geist, Matthieu and Bachem, Olivier},
 booktitle = {International Conference on Learning Representations},
 editor = {B. Kim and Y. Yue and S. Chaudhuri and K. Fragkiadaki and M. Khan and Y. Sun},
 pages = {21246--21263},
 title = {On-Policy Distillation of Language Models: Learning from Self-Generated Mistakes},
 url = {https://proceedings.iclr.cc/paper_files/paper/2024/file/5be69a584901a26c521c2b51e40a4c20-Paper-Conference.pdf},
 volume = {2024},
 year = {2024}
}

@misc{li2025naturalthoughtsselectingdistillingreasoning,
      title={{NaturalThoughts}: Selecting and Distilling Reasoning Traces for General Reasoning Tasks}, 
      author={Yang Li and Youssef Emad and Karthik Padthe and Jack Lanchantin and Weizhe Yuan and Thao Nguyen and Jason Weston and Shang-Wen Li and Dong Wang and Ilia Kulikov and Xian Li},
      year={2025},
      eprint={2507.01921},
      archivePrefix={arXiv},
      primaryClass={cs.CL},
      url={https://arxiv.org/abs/2507.01921}, 
}

@InProceedings{pmlr-v267-chu25c,
  title = 	 {{SFT} Memorizes, {RL} Generalizes: A Comparative Study of Foundation Model Post-training},
  author =       {Chu, Tianzhe and Zhai, Yuexiang and Yang, Jihan and Tong, Shengbang and Xie, Saining and Schuurmans, Dale and Le, Quoc V and Levine, Sergey and Ma, Yi},
  booktitle = 	 {Proceedings of the 42nd International Conference on Machine Learning},
  pages = 	 {10818--10838},
  year = 	 {2025},
  editor = 	 {Singh, Aarti and Fazel, Maryam and Hsu, Daniel and Lacoste-Julien, Simon and Berkenkamp, Felix and Maharaj, Tegan and Wagstaff, Kiri and Zhu, Jerry},
  volume = 	 {267},
  series = 	 {Proceedings of Machine Learning Research},
  month = 	 {13--19 Jul},
  publisher =    {PMLR},
  url = 	 {https://proceedings.mlr.press/v267/chu25c.html}
}

@misc{luo2026agentarkdistillingmultiagentintelligence,
      title={{AgentArk}: Distilling Multi-Agent Intelligence into a Single {LLM} Agent}, 
      author={Yinyi Luo and Yiqiao Jin and Weichen Yu and Mengqi Zhang and Srijan Kumar and Xiaoxiao Li and Weijie Xu and Xin Chen and Jindong Wang},
      year={2026},
      eprint={2602.03955},
      archivePrefix={arXiv},
      primaryClass={cs.AI},
      url={https://arxiv.org/abs/2602.03955}, 
}

@misc{amani2026rlreasoningadaptivelyrevealing,
      title={{RL} for Reasoning by Adaptively Revealing Rationales}, 
      author={Mohammad Hossein Amani and Aryo Lotfi and Nicolas Mario Baldwin and Samy Bengio and Mehrdad Farajtabar and Emmanuel Abbe and Robert West},
      year={2026},
      eprint={2506.18110},
      archivePrefix={arXiv},
      primaryClass={cs.LG},
      url={https://arxiv.org/abs/2506.18110}, 
}

@misc{jiang2026drpdistilledreasoningpruning,
      title={{DRP}: Distilled Reasoning Pruning with Skill-aware Step Decomposition for Efficient Large Reasoning Models}, 
      author={Yuxuan Jiang and Dawei Li and Francis Ferraro},
      year={2026},
      eprint={2505.13975},
      archivePrefix={arXiv},
      primaryClass={cs.CL},
      url={https://arxiv.org/abs/2505.13975}, 
}
